\documentclass{elsarticle}

\journal{Pattern Recognition}

\usepackage{color}
\usepackage{soul}
\usepackage{array}
\usepackage{float}
\usepackage{units}
\usepackage{url}
\usepackage{multirow, bigdelim}
\usepackage{amsthm}
\usepackage{amsmath}
\usepackage{amssymb}
\usepackage{xfrac}
\usepackage{stackrel}
\usepackage{graphicx}
\usepackage{setspace}
\usepackage{esint}
\usepackage{makecell}
\usepackage[colorlinks=true,linkcolor=black, citecolor=blue, urlcolor=blue]{hyperref}
\usepackage{epstopdf} 
\usepackage{algorithm}
\usepackage[noend]{algpseudocode}
\usepackage{natbib}
\makeatletter

\providecommand{\tabularnewline}{\\}
\floatstyle{ruled}
\newfloat{algorithm}{tbp}{loa}
\providecommand{\algorithmname}{Algorithm}
\floatname{algorithm}{\protect\algorithmname}

\usepackage[caption=true,font=normalsize,labelfont=sf,textfont=sf,margin=10pt,indention=10pt,singlelinecheck=false]{subfig}
\usepackage{relsize}
\usepackage[table]{xcolor}
\usepackage{diagbox, pict2e}

\usepackage{setspace}
\allowdisplaybreaks

\makeatletter
\newcommand{\thickhline}{%
	\noalign {\ifnum 0=`}\fi \hrule height 3.5pt
	\futurelet \reserved@a \@xhline
}
\newcolumntype{I}{@{\hskip\tabcolsep\vrule width 3.5pt\hskip\tabcolsep}}

\def\ps@pprintTitle{%
	\let\@oddhead\@empty
	\let\@evenhead\@empty
	\def\@oddfoot{\reset@font\hfil\thepage\hfil}
	\let\@evenfoot\@oddfoot
}

\makeatother

\begin{document}
\begin{frontmatter}
\title{Hierarchical mixture of discriminative Generalized Dirichlet classifiers}

\author [UdS]{Elvis Togban\corref{cor1}}
\ead{E.Togban@usherbrooke.ca}

\author [UdS]{Djemel Ziou\corref{cor2}}
\ead{Djemel.Ziou@usherbrooke.ca}
\address[UdS]{ D\'{e}partement d'Informatique, Facult\'{e} des Sciences, 2500 Bl. de l'universit\'{e}, Universit\'{e} de Sherbrooke, J1K2R1, Sherbrooke, Qu\'{e}bec, Canada.}

\begin{abstract}
	This paper presents a discriminative  classifier for compositional data. This classifier is based on the posterior distribution of the Generalized Dirichlet which is the discriminative counterpart of Generalized Dirichlet mixture model. Moreover, following the mixture of experts paradigm, we proposed a hierarchical mixture of this classifier. In order to learn the models parameters, we use a variational approximation by deriving an upper-bound for the Generalized Dirichlet mixture. To the best of our knownledge, this is the first time this bound is proposed in the literature. Experimental results are presented for spam detection and color space identification.
	
\end{abstract}


\end{frontmatter}

\section{Introduction \label{sec:introduction}}
The massive growth of digital applications leads to the use of several types of data which need to be modeled for their description or classification. For the latter one task, both generative and discriminative models can be applied. While generative models rely on the distribution that generates the attributes to classify an object \cite{bouguila2006hybrid, fan2021unsupervised}, the discriminative models focus on determining the class boundaries \cite{greenacre2018compositional, togban2018classification}. In the case where the distribution or the data is correctly chosen, generative models can outperform discriminative ones. However, in a real-world context, it is very difficult to find the best distribution that fit the data and then discriminative models are preferred.  Some works try to combine discriminative and generative models together in order to get the best from both approaches \cite{bouchard2004tradeoff,bernardo2007generative,masoudimansour2016generalized}. In this work, we will focus on the discriminative models since they often produce more accurate classifiers.

  Compositional data are present in fields such as ecology, chemical, economics, data mining and pattern recognition \cite{greenacre2018compositional}. Examples include molarities, percentage of income and  histograms. Compositional data are multidimensional, bounded, positive, summing up to constant and lie in a simplex. Unsupervised generative models have been used extensively to classify such data \cite{bouguila2004unsupervised,bouguila2006hybrid,fan2017online,fan2017proportional}. In the context of discriminative learning, previous works adopt two approaches to the classification of compositional data. The first one uses a preprocessing step to obtain unconstrained data \cite{aitchison1982statistical,Tsagris2016, greenacre2018compositional}. Then, a standard discriminative model is applied to the transformed data. A general preprocessing step  named  $\alpha$-transformation has the following form \cite{Tsagris2016}: 

\begin{equation}
	\mathbf{z}_{\alpha}(x)=\mathbf{H} \cdot\left(\frac{(D+1) \mathbf{u}_{\alpha}(x)-\mathbf{1}_{(D+1)}}{\alpha}\right),
	\label{eq:alpha-transf}
\end{equation}
where $\alpha > 0$, $\mathbf{1}_{(D+1)}$ is an unit vector  and $x$ is a compositional vector. Both,  $\mathbf{1}_{(D+1)}$ and $x$  have $D+1$ dimensions. The  matrix $\mathbf{H}$ is obtained by deleting the first row of the Helmert matrix \cite{lancaster1965helmert} and 

\begin{equation}
	\mathbf{u}_{\alpha}(x)=\left(\frac{x_{1}^{\alpha}}{\sum_{d=1}^{D+1} x_{d}^{\alpha}}, \cdots, \frac{x_{D+1}^{\alpha}}{\sum_{d=1}^{D+1} x_{d}^{\alpha}}\right)^{T}
	\label{eq:alpha-power}
\end{equation}
is the power-transformation \cite{aitchison1982statistical}. In the case where  $\alpha \rightarrow 0$, we can retrieve two others transformations named, isometric log-ratio transform  \cite{egozcue2003isometric} and centered log-ratio transform  \cite{aitchison1982statistical}.
Once the $\alpha$-transformation is performed, the data obtained lie outside the simplex. In the second approach, authors  build  adequate  kernels and apply a standard model like SVM to classify the data \cite{bouguila2012hybrid,bouguila2013deriving,bourouis2018deriving}.  However, from both approaches, the resulting models are only explainable in the transformed space and have no straightforward meaning. For example, the data obtained from the kernel trick are composed of similarities measure between pairs of data. These measures don't give any information about the initial features. Moreover, the derivation of  adequate kernels for the compositional data has a high computational cost \cite{bourouis2018deriving}. In fact, a preliminary step to the construction of these kernels is the modeling of each data instance by a generative mixture model.

The aim of this paper is to address the problem of supervised  compositional data modeling. In contrast to the previous works, we proposed an approach where the data remain in the simplex. In this approach, we used the posterior distribution of the Generalized Dirichlet (GD) as classification model. This choice is justified by the fact that Generalized Dirichlet is  suitable for the compositional data modeling \cite{bouguila2010dirichlet,fan2017online,fan2017proportional}. We call our model the \textit{Discriminative Generalized Dirichlet} (\textit{DGD}). This classifier  is the natural discriminative equivalent of the mixture of GD (MGD) which is a generative model. At the best of our knowledge, the posterior distribution of GD has never been applied as a discriminative classifier \footnote{In  a generative context,  a posteriori probability is computed after the estimation of the mixture  parameters. However, in this work, we directly  estimate the parameters of the posteriori.}. This can be due to the fact that the mixture of GD term that appears in the posterior leads to an intractable likelihood. In this paper, we derive an upper-bound for the mixture of GD and then, we go beyond that limitation.  

In presence of a difficult classification task, the proposed model DGD can fail due to the complex relationship between the classes and the attributes. However, it is possible to combine several DGD models through ensemble learning strategies like  bagging  \cite{breiman1996bagging} or  random subspace \cite{Ho1998}, just to name a few. In this paper, we focus on the Hierarchical mixture of experts (HME) paradigm \cite{Jordan1993}. The idea behind HME model is to softly split the data through a set of gating functions. Then, each expert learns from data lying in a specific region. HME can also viewed as a  combination of several classifiers through the gating functions. Usually, both experts and gating functions are chosen as linear models. However, in this paper, the experts and the gating functions are based on DGD model. It is possible to use deep learning methods to deal with compositional data classification, however, in this paper we focus on a statistical approach. To sum up, the goal of this paper is to build  discriminative models for the classification of compositional data. We achieve that  by:
\begin{itemize}
	\item proposing a classifier based on the Generalized Dirichlet distribution
	\item building an hierarchy of this classifier to obtain a meta-classifier
	\item  deriving an upper-bound  for the mixture of GD in order to estimate the parameters
	\item comparing our approach to the existing ones.
\end{itemize}

The paper is structured as follows. In section \ref{Hier_Discr}, we present the Hierarchical mixture of DGD experts. In section \ref{ML_est}, we present an upper-bound for the mixture of Generalized Dirichlet followed by the parameters estimation. In section \ref{Experimental_Results}, we evaluate our models through two real-world applications namely, spam detection and color space identification. 

\section{Hierarchical mixture of discriminative Generalized Dirichlet classifiers \label{Hier_Discr}}
\subsection{The Generalized Dirichlet distribution \label{the_model}}
Let $\overrightarrow{\mathbf{x}}=\{\left(x_{1},\cdots,x_{D+1}\right)^{T}\in\mathbb{R}^{D+1} \mid \;\overset{D+1}{\underset{d=1}{\sum}}x_{d}=A;\;A\geq1\}$ be a random vector following a scaled Generalized Dirichlet distribution with density function:
\begin{equation}
	GD(\mathbf{x}| \mu)= \overset{D}{\underset{d=1}{\mathlarger{\prod}}} \dfrac{x_{d}^{a_{d}-1} [A-\overset{d}{\underset{l=1}{\sum}}x_{l}] ^{\gamma_d}}{A^{(a_{d}+b_{d}-1)}\mathtt{B}(a_{d},b_{d})} 
	\label{eq: GD}
\end{equation}
where $(.)^{T}$ is the transposition operation, $\mathtt{B}(a_{d},b_{d})$ is the Beta function, $0<x_d<A$, $\{a_{d},b_{d}\}>0$, $\gamma_d=b_{d}-(a_{d+1}+b_{d+1})$ for $d=1 \cdots D-1$; $\gamma_D=b_{D}-1$ and $\mu ={a_1,b_1,\cdots,a_D,b_D}$. The Generalized Dirichlet distribution is a natural choice to model  the compositional data  since it is defined on the simplex $ S_{D-1}= \{ (x_{1},\cdots,x_{D}),\;\overset{D}{\underset{d=1}{\sum}}x_{d}<A \}$. Note that the Generalized Dirichlet (GD) distribution  is reduced to a Dirichlet distribution when $b_{d-1}=(a_{d}+b_{d})$. However, the first one has a more general covariance structure.  The GD distribution has been widely used in generative models for compositional data classification \cite{bouguila2010dirichlet,fan2017online}. In the context of discriminative models, the GD distribution has been used to derive kernels functions \cite{bouguila2012hybrid,bouguila2013deriving,bourouis2018deriving}.  Through a mathematical property  of the GD distribution, equation (\ref{eq: GD}) can be expressed in terms of  independent variables \cite{fan2017online}:

\begin{equation}
	\begin{array}{rll}
		GD(\mathbf{x}| \mu)=&  \overset{D}{\underset{d=1}{\mathlarger{\prod}}} \mathtt{Beta}(\mathtt{v}_{d}|a_{d},b_{d})=& \overset{D}{\underset{d=1}{\mathlarger{\prod}}} \dfrac{\mathtt{v}_{d}^{a_{d}-1}(A-\mathtt{v}_{d})^{b_{d}-1}}{A^{(a_{d}+b_{d}-1)}\mathtt{B}(a_{d},b_{d})}
	\end{array}
	\label{eq: GD_in}
\end{equation}
where $\mathtt{Beta}(\mathtt{v}_{d}|a_{d},b_{d})$ is the scaled Beta distribution with parameters $\{a_{d},b_{d}\}$, $\mathtt{v}_{1}=x_{1}$ and $\mathtt{v}_{d}=\mbox{\Large\(\frac{x_{d}}{A- \sum_{l=1}^{d-1}x_{l}}\)} $.

\subsection{Discriminative classifiers based on Generalized Dirichlet\label{relation}}
\begin{figure}[!h]
	
	\begin{centering}
		\hfill{} \includegraphics[width=.8\columnwidth]{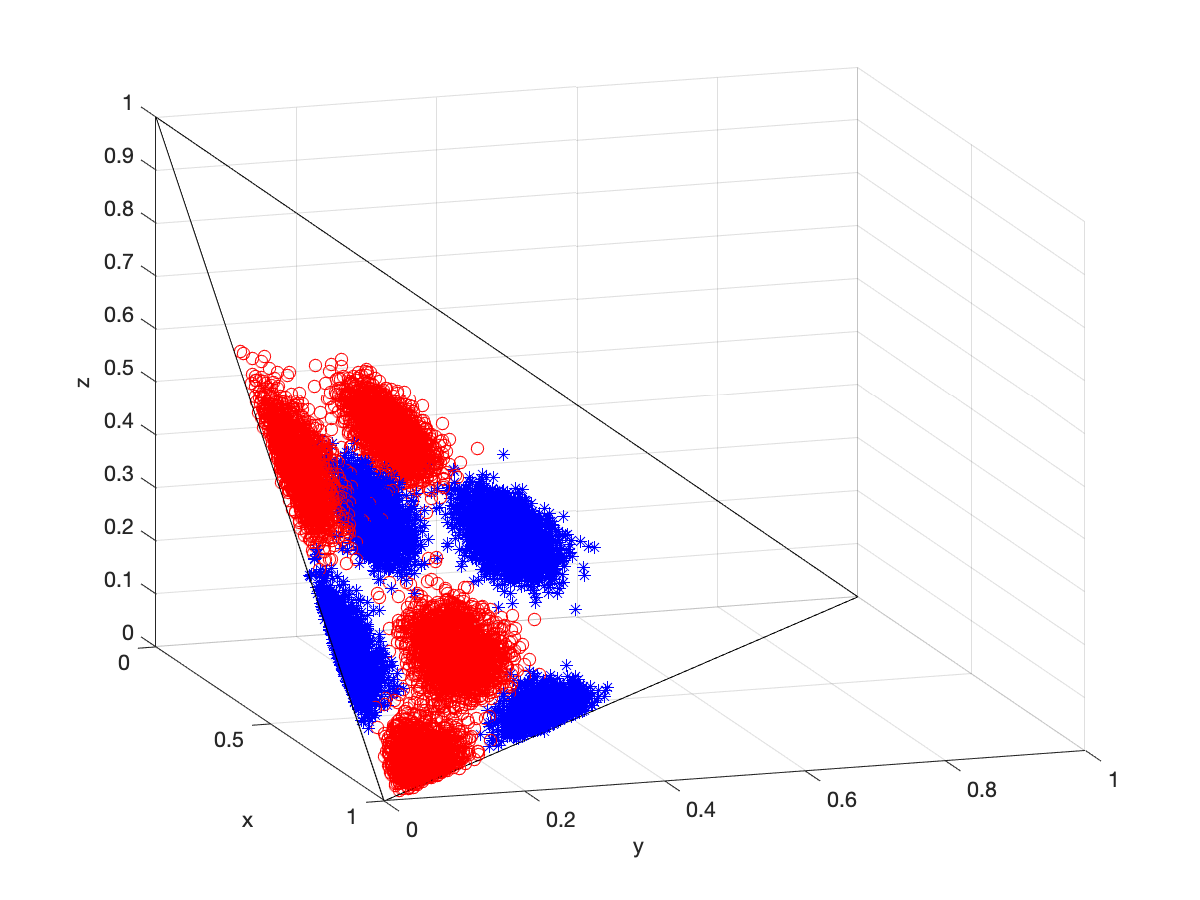}\hfill{}
		
		\protect\caption{Compositional data illustration. \label{fig:Illustr_data_comp}}
	\end{centering}
\end{figure}

A discriminative classifier is defined by mapping the variable $\mathbf{x}$ to a class label $\mathbf{y}$ through a discriminant function that can be a posterior probability $p(\mathbf{y}|\mathbf{x},\Omega)$ or a confident score $Conf(\mathbf{y}|\mathbf{x},\Omega)$ controlled by a set of parameters $\Theta$.  Let $\alpha_{c}$ be the  probability  that  $x$ belongs to the class $c$ ($p(y=c)$). Assume that each class can be described by a GD  distribution: $p(x|\mu_{c})=GD(x|\mu_{c})$. The following \textit{a posteriori} probability  can be used to classify compositional data:

\begin{equation}
	\begin{array}{r}
		p(y=c|x,\,\Omega)=\dfrac{\alpha_{c}GD(x|\mu_{c})}{\overset{C}{\underset{k=1}{\sum}}\alpha_{k}GD(x|\mu_{k})}=\pi_{c} \mid \, \alpha_{k}\geq 0; \overset{C}{\underset{k=1}{\sum}}\alpha_{k}=1.
	\end{array}	
	\label{eq: SGD}
\end{equation}
where $\Omega=\left\{\alpha_{k}, \mu_{k}\right\}_{k=1}^{C}$ represent the classifier parameters, $p(y=c|x,\,\Omega)$ is the probability to belong to the $c^{th}$ class and $C$ is the number of classes. We call this model,  Discriminative Generalized Dirichlet (DGD). Note that the GD distributions estimated from DGD do not necessarily fit the data since we only focus on the classes boundary. In fact, discriminative model eliminate the parameters $\Omega$ that influence only the supposed distribution of the data and  keep those maximizing $p(y=c|x,\,\Omega)$.  Figure \ref{fig:Illustr_data_comp} shows a classification problem  where the compositional data belong to two classes ("o" in red and "$*$" in blue).

\captionsetup[figure]{ width=0.9\columnwidth, justification=centering}

\subsection{Hierarchical mixture of DGD classifiers}
\begin{figure*}[!tbh]
	
	\centering
	\captionsetup[subfigure]{justification=centering}
	\subfloat[Splitting process.  \label{fig: Illustr_hmgd_detail}]{\protect\includegraphics[width=.5\columnwidth]{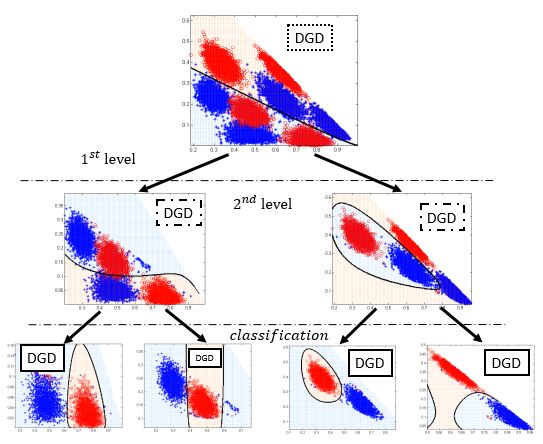}}\hfill{}\subfloat[Final decision boundary. \label{fig: Illustr_hmgd_all}]{\protect\includegraphics[width=0.45\columnwidth]{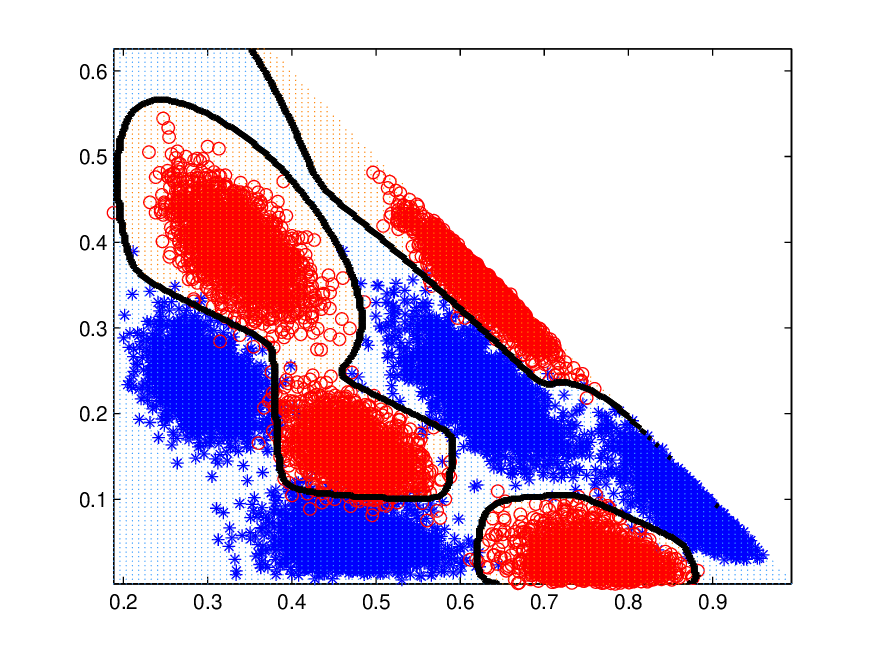}
	}\hfill{}	\protect\caption{Illustration of the HMGD splitting process and the final decision boundary. The solid curve is the regions (resp. sub-regions or classes) boundary. In "a)" each split is performed by a \textit{DGD} model. \label{fig: Illustr_hmgd}}	
\end{figure*}

As described in \cite{togban2018classification}, when we face a difficult classification task, we can hierarchically combine several classifiers to form a meta-classifier. Firstly, at each level of the hierarchy, a set of parametric gating functions split the input space into several regions or modes. Finally, each classifier discriminates the data belonging to a specific region. Intuitively, the gating functions perform a classification task with an unknown label which refers to a region. Since the gating functions and the classifiers perform a classification task on the compositional data, we  can model them with the classifier defined in equation (\ref{eq: SGD}). We call this hierarchy of DGD models, \textit{Hierarchical mixture of DGD experts} (HMGD). This model can be viewed as a tree where each node (root, inner nodes and leaves) is linked to a \textit{DGD} model. The root is related to the first division of the data into regions. The inner nodes are related to the division of the regions (resp. sub-regions) into sub-regions and the leaves embed the base classifiers (see Fig. \ref{fig: Illustr_hmgd}). The base classifiers will be called experts as in HME literature \cite{Jordan1993}. For a tree of two levels, we have the following formulation for HMGD:

\begin{equation}
	\begin{array}{r}
		p\left(y|x,\,\Omega\right)=\overset{K}{\underset{i=1}{\sum}}\pi_{i}^{(0)}\left(x\right)\overset{M_i}{\underset{j=1}{\sum}}\pi_{j|i}^{(1)}\left(x\right)p\left(\mathbf{y}|x,\,\Omega_{ij}^{(2)}\right) \mid 
		\, \pi_{i}^{(0)}\left(x\right)\geq 0;\;\pi_{j|i}^{(1)}\left(x\right)\geq 0
		\\
		s.t.\,\overset{K}{\underset{i=1}{\sum}}\;\pi_{i}^{(0)}\left(x \right)=1;\;\overset{M_i}{\underset{j=1}{\sum}}\pi_{j|i}^{(1)}\left(x\right)=1.
		
	\end{array}\label{eq:Hierar_equation}
\end{equation}
where $\pi_{i}^{(0)}\left(x\right)$ and $\pi_{j|i}^{(1)}\left(x\right)$ are respectively the gating functions of the first and second levels. The number of regions and sub-regions are respectively denoted by $K$ and $M_i$. For a tree with $L$ levels, we denote $\Omega=\{\Omega^{(l)}\}^{L}_{l=0}$ the set of all parameters where   $\Omega^{(l)}=\{\alpha^{(l)},\mu^{(l)}\}^{L}_{l=0}$
is the set of parameters at the $l^{th}$ level. The   gating functions parameters are $\Omega^{(0)}$ and $\{\Omega^{(l)}\}^{L-1}_{l=1}$  and the experts parameters are $\Omega^{(L)}$. 
The model \textit{HMGD} is an extension of the model \textit{DME} that we proposed in an earlier work \cite{Togban2017}. There is no hierarchy in \textit{DME} and the gating functions are based on the Dirichlet distribution. Moreover, the experts used are multinomial logistic regression.

\captionsetup[figure]{ width=.9\columnwidth}

\section{Parameters Estimation using Maximum Likelihood (ML) \label{ML_est}}
Let $\mathbf{Obs}=\left\{ \mathbf{X},\,\mathbf{Y}\right\} $ be
a set of observed data where $\mathbf{X}=\{x_{1},\cdots,x_{N}\}\in\mathbb{R}^{D\times N}$
is a set of i.i.d. vectors and $\mathbf{Y}=\left(y_{1},\cdots,y_{N}\right)^{T}\in\mathbb{R}^{N}$
their equivalent class labels; $N$ is the number of observations.  Since our model (eq. \ref{eq:Hierar_equation}) is a mixture model, we can use EM algorithm for the training process. As in \cite{togban2018classification}, we introduce a  set of `\textit{hidden}' binary random variables for the $n^{th}$ data: $z_{n,i}^{(0)}$ and $z_{n,j|i}^{(1)}$ respectively for the top-level and the lower-level of the tree.  If $x_{n}$ belongs to the region $i$ then $z_{n,i}^{(0)}=1$ and 0, otherwise. Given the region $i$, if $x_{n}$ belongs to the sub-region $j$ then $z_{n,j|i}^{(1)}=1$ and 0, otherwise. The \textit{complete-log-likelihood} can be written as follows: 

\begin{equation}
	\begin{array}{rr}
		\mathcal{L}^{c}= & \overset{N}{\underset{n=1}{\sum}}\overset{K}{\underset{i=1}{\sum}}z_{n,i}^{(0)}\overset{M_i}{\underset{j=1}{\sum}}z_{n,j|i}^{(1)}\ln\left(\pi_{i}^{(0)}\left(x_{n}\right)\pi_{j|i}^{(1)}\left(x_{n}\right) p\left(y_{n}|x_{n},\,\Omega_{ij}^{(2)}\right)\right)
	\end{array}
\end{equation}	
In the E-step we compute the following expectation:
\begin{equation}
	\begin{array}{r}
		\Phi\left(\Omega,\,\Omega^{t}\right)=\overset{N}{\underset{n=1}{\sum}}\overset{K}{\underset{i=1}{\sum}}h_{n,i}^{(0)}\overset{M_i}{\underset{j=1}{\sum}}h_{n,j|i}^{(1)}\ln\left(\pi_{i}^{(0)}\left(x_{n}\right)\pi_{j|i}^{(1)}\left(x_{n}\right)p\left(y_{n}|x_{n},\,\Omega_{ij}^{(2)}\right)\right)
	\end{array}\label{eq:Hierar_mixt_Expert_stan}
\end{equation}
where \textit{$\Omega^{t}$} is the parameters at the $t^{th}$ iteration; $h_{n,i}^{(1)}$ and  $h_{n,j|i}^{(2)}$  are respectively the expected values of $z_{n,i}^{(0)}$ and $z_{n,j|i}^{(1)}$  and their expressions are given by
\begin{equation}
	\begin{array}{ll}
		h_{n,i}^{(0)}  = & \mbox{\Large\( \frac{\pi_{i}^{(0)}\left(x_{n}\right)\overset{M_i}{\underset{j=1}{\sum}}\pi_{j|i}^{(1)}\left(x_{n}\right)p\left(y_{n}|x_{n},\,\Omega_{ij}^{(2)}\right)}{\overset{K}{\underset{i=1}{\sum}}\pi_{i}^{(0)}\left(x_{n}\right)\overset{M_i}{\underset{j=1}{\sum}}\pi_{j|i}^{(1)}\left(x_{n}\right)p\left(y_{n}|x_{n},\,\Omega_{ij}^{(2)}\right)} \)}
	\end{array} \label{eq: respons_top}
\end{equation}
and
\begin{equation}
	\begin{array}{ll}
		h_{n,j|i}^{(1)} = & \mbox{\Large\( \frac{\pi_{j|i}^{(1)}\left(x_{n}\right)p\left(y_{n}|x_{n},\,\Omega_{ij}^{(2)}\right)}{\overset{M_i}{\underset{j=1}{\sum}}\pi_{j|i}^{(1)}\left(x_{n}\right)p\left(y_{n}|x_{n},\,\Omega_{ij}^{(2)}\right)} \)}
	\end{array} \label{eq: respons_iner}
\end{equation}

In the M-Step, we have to perform the following  maximization operations:

\begin{equation}
	\begin{array}{r}
		\Omega_{(t+1)}^{(0)}=argmax_{\Omega^{^{(0)}}}\overset{N}{\underset{n=1}{\sum}}\overset{K}{\underset{i=1}{\sum}}h_{n,i}^{(0)}\ln\left(\pi_{i}^{(0)}\left(x_{n}\right)\right)\end{array}\label{eq:first_layer_max}
\end{equation}
\begin{equation}
	\begin{array}{lr}
		\Omega_{i,(t+1)}^{(1)}= & argmax_{\Omega_{i}^{^{(1)}}}\overset{N}{\underset{n=1}{\sum}}h_{n,i}^{(0)}\overset{M_i}{\underset{j=1}{\sum}}h_{n,j|i}^{(1)}\ln\left(\pi_{j|i}^{(1)}\left(x_{n}\right)\right)
	\end{array}\label{eq:second_layer_max}
\end{equation}
\begin{equation}
	\begin{array}{r}
		\Omega_{ij,(t+1)}^{(2)}=argmax_{\Omega_{ij}^{^{(2)}} }\overset{N}{\underset{n=1}{\sum}} h_{n,i}^{(0)} h_{n,j|i}^{(1)}\overset{C}{\underset{c=1}{\sum}} v_{n,c}  \ln\left(p\left(y_{n}=c|x_{n},\,\Omega_{ij}^{(2)}\right)\right)\end{array}\label{eq:expert_layer_max}
\end{equation}
where $\Omega_{(t+1)}^{(l)}$  is  the parameters $\Omega^{(l)}$ at the $(t+1)^{th}$ iteration and $v_{n,c}$ are binary variables indicating membership of $n^{th}$
data to the $c^{th}$ class. In general, for each node (at the $l^{th}$ level) being the $i^{th}$ child of its parent, we have to solve a weighted-DGD:

\begin{equation}
	\begin{array}{l}
		\Omega_{i,(t+1)}^{(l)}= argmax_{\Omega_{i}^{^{(l)}}}\Phi^{(l)} \mid
		\Phi^{(l)}= \overset{N}{\underset{n=1}{\sum}}H_{n,i}^{(l-1)}\overset{M_i}{\underset{j=1}{\sum}}h_{n,j|i}^{(l)}\ln\left(\pi_{j|i}^{(l)}\left(x_{n}\right)\right)
	\end{array}\label{eq:general_layer_max}
\end{equation}
where $h_{n,j|i}^{(l)}$ acts like a label, $H_{n,i}^{(l-1)}$ is the product of all responsibilities along the path going from the root to the current node and act like a weight. When we are in the presence of a leaf, $h_{n,j|i}^{(L)}\equiv v_{n,c}$ and $M_{i}\equiv C$. As shown in figure (\ref{fig: Illustr_hmgd_detail}), the plot at the first level is obtained by performing a \textit{DGD} model with a "virtual label" while at the second level, we perform a weighted-\textit{DGD} with a "virtual label". At the classification step, we perform a weighted-\textit{DGD} with the true label.

\subsection{Upper-bound for the GD mixture model \label{sub: upper-bound}}
\begin{figure}[!tbh]
	
	\centering
	\captionsetup[subfigure]{justification=centering}
	
	\subfloat[Log-Likelihood to bound.  \label{fig: lli_illustr}]{\protect\includegraphics[width=0.49\columnwidth]{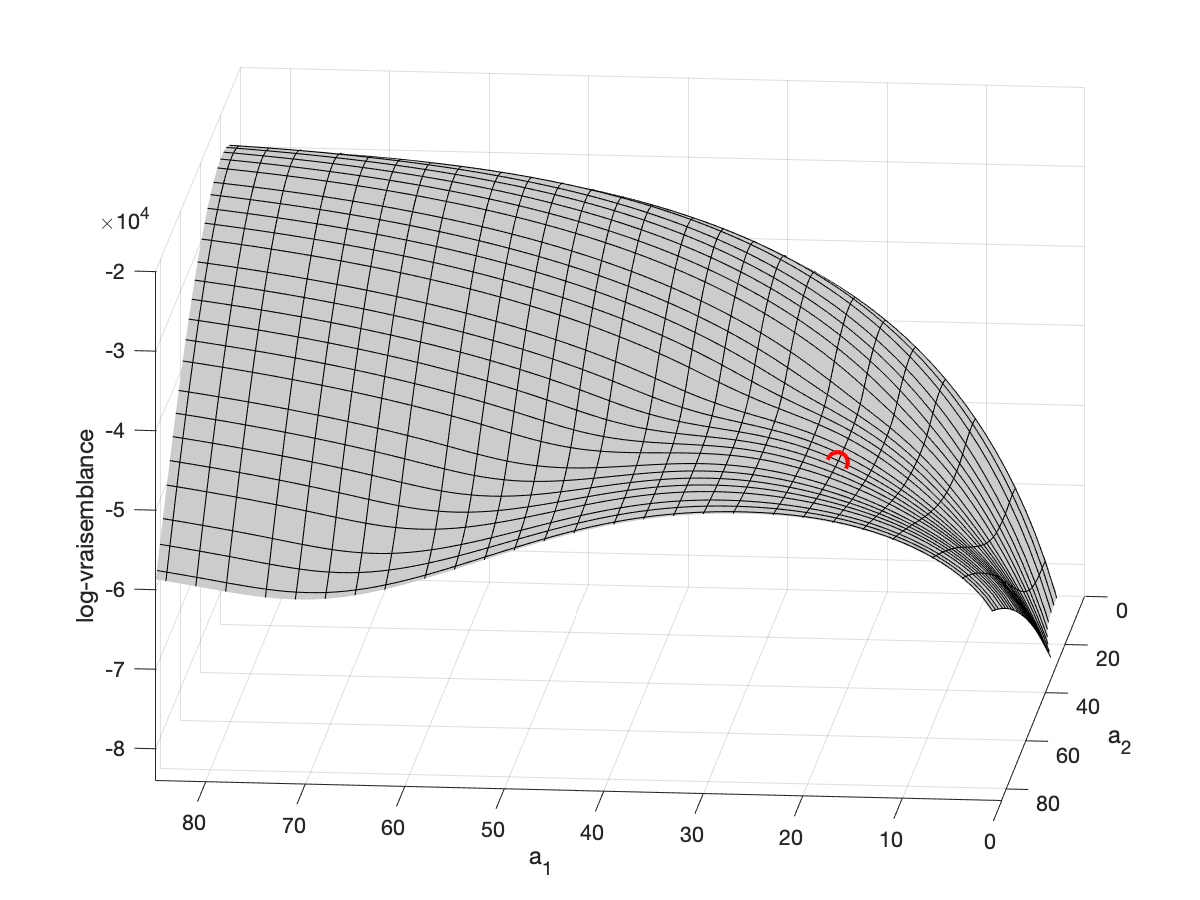}}\hfill{}\subfloat[Upper Bound of the Log-Likelihood. \label{fig: lli_upp}]{\protect\includegraphics[width=0.49\columnwidth]{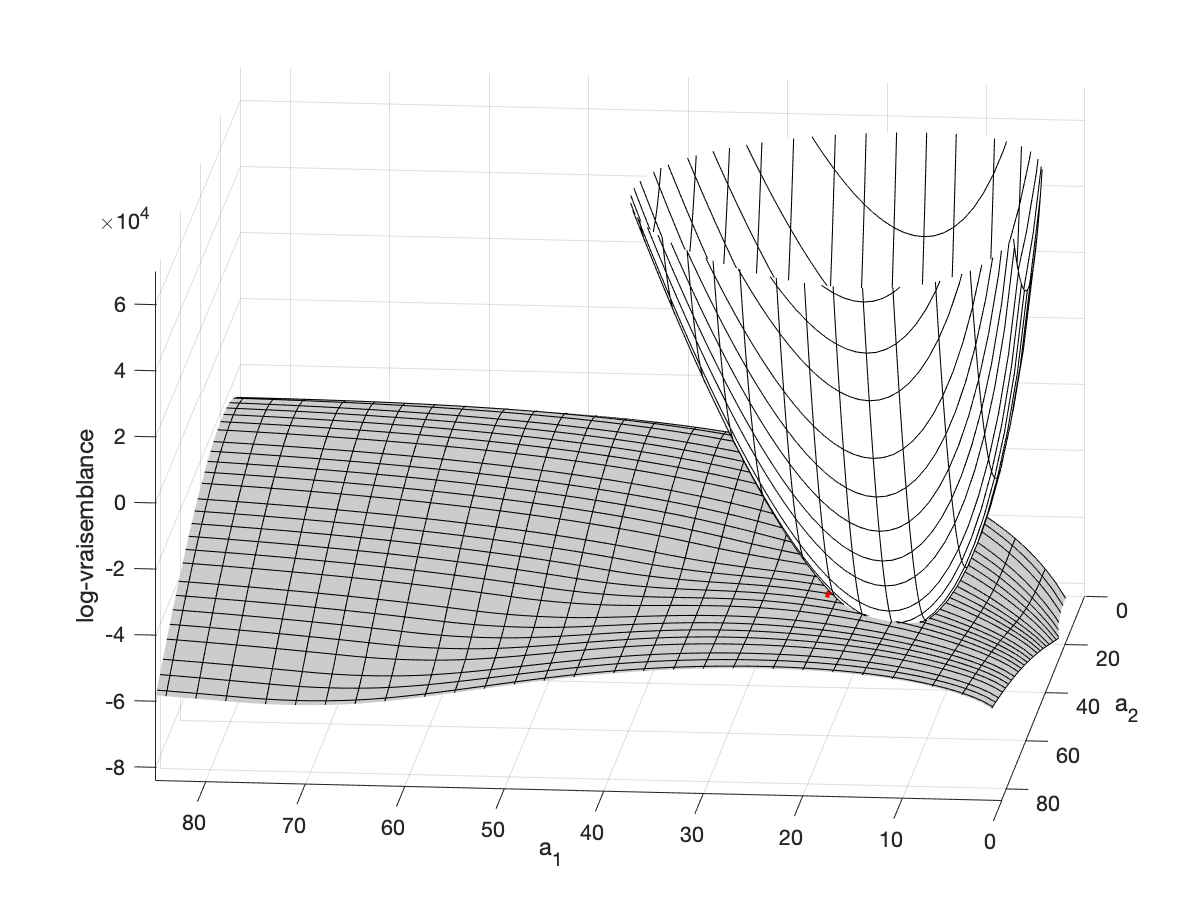}
	}\hfill{}
	
	\protect\caption{Log-Likelihood of two Beta mixture and  its upper-bound generated at the red circle point.\label{fig:mixture_beta_bound}}
\end{figure}

Given the expression of \textit{DGD} model (eq. \ref{eq: SGD}), the maximization operations in the M-Step  lead to highly nonlinear optimization problems. The term leading to an optimization challenge is the mixture appearing in the denominator of equation (\ref{eq: SGD}). Previous works \cite{xu1995alternative,Togban2017,togban2018classification} deal with this problem by performing  the EM algorithm  on the joint probability $ p(\mathbf{y},\mathbf{x}|\,\Omega)$. Unfortunately, this solution is not suitable for a hierarchical case since it simplifies the problem just for the first level parameter. However, it is possible to circumvent this problem by using a variational  approximation. Therefore, we have to optimize a lower-bound of the objective function $\Phi^{(l)}$ described in  equation \ref{eq:general_layer_max}. This is equivalent to determine a lower-bound for the term  $\ln\left( \dfrac{\alpha_{ij}^{(l)}GD(x|\mu_{ij}^{(l)})}{\underset{k=1}{\sum}\alpha_{ik}^{(l)}GD(x|\mu_{ik}^{(l)})}\right)$. Specifically, we need to find an upper-bound for $\ln\left(\underset{k}{\sum}\alpha_{ik}^{(l)}GD(x|\mu_{ik}^{(l)})\right)$. Exploiting the reverse Jensen inequality for a mixture of exponential family proposed in \cite{jebara2001discriminative}, we derived the following bound: 
\begin{equation}
	\begin{array}{l}
		\ln \left( \underset{k}{\sum}\alpha_{ik}^{(l)}GD(x_{n}|\mu_{ik}^{(l)}) \right) \leq  \underset{k}{\sum}-W_{n,ik}^{(l)} \left[\ddot{\mathbf{x}}_{n,ik}^{(l)T}\bar{\Omega}_{ik}^{(l)}-\mathcal{K}(\bar{\Omega}_{ik}^{(l)})\right] + cst_{n,i}
	\end{array} \label{eq: upper-bound}
\end{equation}
Where
\begin{align}
	cst_{n,i}=\ln \left( \underset{k}{\sum}\breve{\alpha}_{ik}^{(l)}GD(x_{n}|\breve{\mu}_{ik}^{(l)}) \right)+ \underset{k}{\sum}W_{n,ik}^{(l)} \left[\ddot{\mathbf{x}}_{n,ik}^{(l)T}\breve{\Omega}_{ik}^{(l)}-\mathcal{K}(\breve{\Omega}_{ik}^{(l)})\right]\nonumber \\
	\ddot{\mathbf{x}}_{n,ik}^{(l)}=\dfrac{\tilde{\pi}_{k|i}^{(l)}\left(x_{n}\right)}{W_{n,ik}^{(l)}}\left[ \mathcal{K}^{\prime}(\breve{\Omega}_{ik}^{(l)})- \bar{\mathbf{x}}_{n,ik}^{(l)} \right] + \mathcal{K}^{\prime}(\breve{\Omega}_{ik}^{(l)})  \label{eq:upper_bound_details} \\
	W_{n,ik}^{(l)}=4G\left( \breve{\pi}_{k|i}^{(l)}\left(x_{n}\right)/2 \right) \left[ \mathcal{Z}_{n,ik}^{(l)t} \mathcal{Z}_{n,ik}^{(l)} \right]
	+ w_{n,ik}^{(l)} \nonumber \\
	\mathcal{Z}_{n,ik}^{(l)}=\mathcal{K}^{\prime \prime}(\breve{\Omega}_{ik}^{(l)})^{-1/2}\left[\bar{\mathbf{x}}_{n,ik}^{(l)}-\mathcal{K}^{\prime}(\breve{\Omega}_{ik}^{(l)}) \right] \nonumber.
\end{align}
More details about the expressions involved in this upper-bound are given in appendix \ref{appendix_upper}. The parameter $\bar{\Omega}_{ik}^{(l)}$ is a re-parameterization of $\Omega_{ik}^{(l)}$ and $\breve{\pi}_{k|i}^{(l)}\left(x_{n}\right)$ is computed at the tangential contact point $\breve{\Omega}_{ik}^{(l)}$ (previous value of $\Omega_{ik}^{(l)}$). The expressions $\ddot{\mathbf{x}}_{n,ik}^{(l)}$ and $W_{n,ik}^{(l)}$ are variational parameters. The scalar $w_{n,ik}^{(l)}$ is the minimum value of $W_{n,ik}^{(l)}$ which guarantee that $\ddot{\mathbf{x}}_{n,ik}^{(l)}$ belong to the tangent space of $\mathcal{K}(\Omega_{ik}^{(l)})$. The expressions $\mathcal{K}^{\prime}(\Omega_{ik}^{(l)})$ and $\mathcal{K}^{\prime \prime}(\Omega_{ik}^{(l)})$ are respectively the gradient and the Hessian of $\mathcal{K}(\Omega_{ik}^{(l)})$ and $G$ is a linear function defined by a lookup table \cite{jebara2001discriminative}. In order to visualize this upper bound, we plot both the log-likelihood of a mixture of beta distribution (Fig. \ref{fig: lli_illustr}) and its equivalent upper-bound (Fig. \ref{fig: lli_upp}). The mixture used is: $0.3\;\mathtt{Beta}(x|a_1,50)+0.7\;\mathtt{Beta}(x|a_2,100)$.
The log-likelihood is plotted  with respect to $a_1$ and $a_2$. The tangential contact point (red circle) is fixed at $a_1=20$ and $a_2=50$. Let us recall that the Beta distribution is a particular case of the GD distribution.

\captionsetup[figure]{ width=0.9\columnwidth}

Given the upper-bound of GD mixture (eq. \ref{eq: upper-bound}), we can derive the following lower-bound :

\begin{equation}
	\begin{array}{r}

		\ln\left( \dfrac{\alpha_{ij}^{(l)}GD(x|\mu_{ij}^{(l)})}{\underset{k=1}{\sum}\alpha_{ik}^{(l)}GD(x|\mu_{ik}^{(l)})}\right) \geq  \ln\left( \alpha_{ij}^{(l)}GD(x|\mu_{ij}^{(l)})\right) + cst_{n,i} \\ 
		+ \underset{k}{\sum}W_{n,ik}^{(l)} \left[\ddot{\mathbf{x}}_{n,ik}^{(l)T}\bar{\Omega}_{ik}^{(l)}-\mathcal{K}(\bar{\Omega}_{ik}^{(l)})\right]
		
	\end{array}
	\label{eq:lower_bounds_post}
\end{equation}

\subsection{Parameters Estimation for DGD and HMGD models}

Given the objective function $\Phi^{(l)}$ (Eq.   \ref{eq:general_layer_max}) and the bound described in equation \ref{eq:lower_bounds_post},  we can perform the maximization operations on the following lower-bound:

\begin{equation}
	\begin{array}{c}
		\Phi_{1}^{(l)}= \underset{n,j}{\sum}H_{n,i}^{(l-1)}\left[h_{n,j|i}^{(l)}\ln\left(\alpha_{ij}^{(l)}GD(x_{n}|\mu_{ij}^{(l)})\right)  \right. \\ \left. + W_{n,ij}^{(l)} \left[\ddot{\mathbf{x}}_{n,ij}^{(l)t}\bar{\Omega}_{ij}^{(l)}-\mathcal{K}(\bar{\Omega}_{ij}^{(l)})\right] + cst_{n,i}\right]\\
		\Phi_{1}^{(l)} \leq \Phi^{(l)}
	\end{array} \label{eq: low-bound}
\end{equation}

Thanks to the transformation of equation \ref{eq: GD_in}, the parameters $\mu_{ij,d}^{(l)}$ can be estimated independently for each $d\in \left\lbrace1\cdots D\right\rbrace$ via the Newton-Raphson method. Since $\mu_{ij,d}^{(l)}$ has to be strictly positive, we re-parametrize by setting $\mu_{ij,d}^{(l)}=e^{\xi_{ij,d}^{(l)}}$ where $\mu_{ij,d}^{(l)}$ is a vector with $a_{ij,d}^{(l)}$ and $b_{ij,d}^{(l)}$ as elements.  The partial derivative of $\Phi_{1}^{(l)}$ (Eq. \ref{eq: low-bound}) with respect to $\xi_{ij,d}^{(l)}$ is given by:
\begin{equation}
	\begin{array}{c}
		\dfrac{\partial \; \Phi_{1}^{(l)}}{\partial\xi_{ij,d}^{(l)}}=\mu_{ij,d}^{(l)} [\mathsf{\Psi} (|\mu_{ij,d}^{(l)}|) -\mathsf{\Psi} (\mu_{ij,d}^{(l)})-\ln(A)] \underset{n}{\sum} H_{n,i}^{(l-1)}   \left[h_{n,j|i}^{(l)}+W_{n,ij}^{(l)}\right]+ \\ \mu_{ij,d}^{(l)}\underset{n}{\sum}H_{n,i}^{(l-1)} \left[h_{n,j|i}^{(l)}\tilde{\mathtt{v}}_{n,d}+W_{n,ij}^{(l)}\chi_{n,ij}^{(l)} \right]
	\end{array} \label{eq: deriv-eps}
\end{equation}
where $\mathsf{\Psi}$ is the Digamma function, $\tilde{\mathtt{v}}_{n,d}=(\ln(\mathtt{v}_{n,d}),\; \ln(A-\mathtt{v}_{n,d}))^{T}$ and $ \displaystyle \chi_{n,ij}^{(l)}=(\ddot{\mathbf{x}}_{n,ij,2d-1}^{(l)},\ddot{\mathbf{x}}_{n,ij,2d}^{(l)})^{T}$. The Hessian $\mathcal{H}_{ij,d}^{(l)}$ is a $2\times2$ symmetric matrix where the diagonal elements are given by:
\begin{equation*}
	\begin{array}{r}
		[\mu_{ij,d}^{(l)}]^{2} \big[\mathsf{\Psi}' (|\mu_{ij,d}^{(l)}|) -\mathsf{\Psi}' (\mu_{ij,d}^{(l)}) \big] 
		\underset{n}{\sum}H_{n,i}^{(l-1)}\left[h_{n,j|i}^{(l)}+W_{n,ij}^{(l)}\right]+ \dfrac{\partial \; \Phi_{1}^{(l)}}{\partial\xi_{ij,d}^{(l)}}
	\end{array}
\end{equation*}
and the anti-diagonal elements are given by
\begin{equation*}
	\begin{array}{l}
		a_{ij,d}^{(l)}b_{ij,d}^{(l)} \mathsf{\Psi}' (|\mu_{ij,d}^{(l)}|) \underset{n=1}{\sum} \left[h_{n,j|i}^{(l)}+W_{n,ij}^{(l)}\right]
	\end{array}
\end{equation*}
where $\mathsf{\Psi}'$ is the Trigamma function and $|.|$ is the $l1-norm$.  Giving an initial $[\xi_{ij,d}^{(l)}]^{old}$, we have the following update equation:
\begin{equation}
	[\xi_{ij,d}^{(l)}]^{new}=[\xi_{id}^{(l)}]^{old}-[\mathcal{H}_{ij,d}^{(l)}]^{-1}\dfrac{\partial \; \Phi_{1}^{(l)}}{\partial\xi_{ij,d}^{(l)}}
\end{equation}

We give more details in appendix \ref{appendix_inv} to compute the inverse of the Hessian $\mathcal{H}_{ij,d}^{(l)}$ in an efficient way. Deriving $\Phi_{1}^{(l)}$ (Eq. \ref{eq: low-bound}) with respect to $\alpha_{ij}^{(l)}$ and taking into account the constraint ${\underset{j}{\sum}}\alpha_{ij}^{(l)}=1$, we have the following update equation, \newline for $j=1\cdots M_i-1$:
\begin{equation}
	\begin{array}{l}

		[\alpha_{ij}^{(l)}]^{new}= \mbox{\Large\( \frac{\underset{n}{\sum}H_{n,i}^{(l-1)}h_{n,j|i}^{(l)}+\underset{n,k}{\sum}H_{n,i}^{(l-1)}W_{n,ik}^{(l)}\ddot{\mathbf{x}}_{n,ik,2D+j}^{(l)}}{\underset{n}{\sum}H_{n,i}^{(l-1)}+\underset{n,k}{\sum}H_{n,i}^{(l-1)}W_{n,ik}^{(l)}}\)} \\
		
		[\alpha_{iM_i}^{(l)}]^{new}=1-\overset{M_i-1}{\underset{j=1}{\sum}}[\alpha_{ij}^{(l)}]^{new}
	\end{array}\label{alph_upd}
\end{equation}
Table \ref{tab: Symb_mean} summarize the main parameters used in our model at the $l^{th}$  level of the tree.To choose the structure of the \textit{HMGD} tree, we can set a large value for $K$ and $M_i$ and remove the components that have a weak value of $\alpha_{ij}^{(l)}$ after each iteration or after the algorithm had converged.

\begin{table}[!htb]
	
		\protect\caption{The main parameters used in our model at the $l^{th}$  level of the tree given the parent node $i$ and the child $j$. \label{tab: Symb_mean}}
	\captionsetup{width=1\columnwidth}
	\centering
	\begin{tabular}{|l||p{0.5\columnwidth}|}
		\hline
		Symbol & Meaning \tabularnewline \hline \hline
		$\alpha_{ij}^{(l)}$ & Mixing coefficients\tabularnewline \hline
		$\mu_{ij}^{(l)}=\left\{a_{ij}^{(l)},b_{ij}^{(l)}\right\}$ & Parameters of the GD distribution\tabularnewline \hline
		
		$\Omega_{ij}^{(l)}=\left\{\mu_{ij}^{(l)},\alpha_{ij}^{(l)}\right\}$ & the parameters of $\pi_{j|i}^{(l)}$ \tabularnewline \hline
		$\mathcal{K}(.)$ & Cumulant function of GD distribution\tabularnewline \hline
		$ W_{n,ij}^{(l)}, \ddot{\mathbf{x}}_{n,ij}^{(l)}$ & Variational parameters for the mixture of GD's upper-bound \tabularnewline \hline
		$\ddot{\mathbf{x}}_{n,ij,d}^{(l)}$& $d^{th}$ element of $\ddot{\mathbf{x}}_{n,ij}^{(l)}$	
		\tabularnewline \hline
	\end{tabular}
\end{table}

The learning procedure for the \textit{DGD} model is  summarized in algorithm \ref{alg:DGD}.

\begin{algorithm}[tbh]
	
	1. Initialization: 
	
	\hspace*{0.25cm}(a) For each class $c$, initialize  $\mu_{c}$
	through the moment method \cite{minka2000estimating}.
	
	\hspace*{0.25cm}(b) set $\left\{\alpha_{c}\right\} _{c=1}^{C}$ as the prior of each class ($\alpha_{c}=N_c/N$) where $N_c$ is the number of instance belonging to the $c^{th}$ class.
	\vspace{0.1cm}
	
	2. E-step: Compute the variational parameters via equation \ref{eq:upper_bound_details} and the \textit{log-likelihood} via equation (\ref{eq:general_layer_max}) by replacing $h_{n,j|i}^{(l)}$ with the true label. Here, the weight $H_{n,i}^{(l-1)}$ is removed.
	
	\vspace{0.1cm}
	
	3. M-step: Maximize $\Phi^{(0)}$ (eq. \ref{eq:general_layer_max}) with respect
	to the current value of $\Omega^{(0)}$ and update $\Omega^{(0)}$ via equations (\ref{eq: deriv-eps}\rule[0.5ex]{0.15cm}{2pt}\ref{alph_upd}).
	\vspace{0.1cm}
	
	4. Repeat steps 2 and 3 until convergence.
	\protect\caption{ Learning algorithm for the \textit{DGD} model \label{alg:DGD}}
\end{algorithm}

Since The \textit{HMGD} is a combination of several (weighted) \textit{DGD}, we can use the learning procedure summarized in algorithm \ref{alg:HMGD}. 

\begin{algorithm}[!tbh]
	
	1. Initialization: 
	
	\hspace*{0.25cm}(a) set the parameters $\left\{ \Omega^{(l)} \right\} _{l=0}^{L-1}$ by using the procedures described in \cite{bouguila2004unsupervised}.
	
	\hspace*{0.25cm}(b) set $\Omega^{(L)}$ like in algorithm (\ref{alg:DGD}) for the instances belonging to each subregions.
	\vspace{0.25cm}
	
	2. E-step: Update $Q\left(\Omega,\,\Omega^{t}\right)$ via equations
	(\ref{eq:Hierar_mixt_Expert_stan}\rule[0.5ex]{0.15cm}{2pt}\ref{eq: respons_iner}) and compute the variational parameters via equation \ref{eq:upper_bound_details}. 
	\vspace{0.25cm}
	
	3. M-step:  the new parameters given the old ones as initialization are obtained for:
	
	\hspace*{0.25cm}(a)  $ \{\pi_{i}^{(0)}\}^{K}_{i=1}$ by running the DGD algorithm with $ \{h_{n,i}^{(0)}\}^{K}_{i=1}$ as labels.
	
	\hspace*{0.25cm}(b) $\{\pi_{j|i}^{(l)}\}^{(L-1),M_i}_{l,j=1}$ by running a weighted DGD algorithm with  $h_{n,j|i}^{(l)}$ as label and $H_{n,i}^{(l-1)}$ as weight for a given $i$ and $l$.
	
	\hspace*{0.25cm}(b) each classifier by running a weighted DGD algorithm with  the true label and $H_{n,i}^{(L-1)}$ as weight.
	\vspace{0.25cm}
	
	4. Repeat steps 2 and 3 until convergence.
	\protect\caption{ Learning algorithm for the \textit{HMGD} model \label{alg:HMGD}}
\end{algorithm}

\section{Experimental Results and discussion \label{Experimental_Results}}
To assess the performance of our models, we have firstly conducted our experiment on several datasets from UCI machine learning repository  \cite{Lichman2013}. Secondly, we have considered  some real-world applications namely spam detection and color space identification. In all the experiments, we set the EM convergence tolerance to $10^{-4}$ over 50 maximum iterations for the DGD model. In the case of HMGD, to reduce the overfitting, we set the maximum number of iterations to 10. For the gating functions and the experts, the maximum number of iterations is respectively set to 5 and 30. In general, we observed that a number of iterations greater than those indicated earlier led to overfitting. Since the GD is not defined for zero values, we replace them with a small value ($1e-4$) \cite{martin2003dealing}.

All the results are based on a five stratified cross-validation (5-CV). This choice is justified by the fact that stratified cross-validation helps to improve the generalization ability of a classifier. We used the accuracy and the Matthew's correlation coefficient (MCC) as evaluation criteria for all the models. In the case of imbalanced data, MCC is a better choice than accuracy,  precision, recall and F1-score regardless which class is positive \cite{mcc_article, chicco2020advantages}. This coefficient vary from $-1$ to $1$. When the values are negatives, the classifier perform poorly and when they are close to zero we have a random guess classifier. The best classifiers have an MCC value close to $1$.

\subsection{Evaluation of DGD and HMGD on UCI datasets}
In this section, we conducted our study on  six datasets from UCI machine learning repository \cite{Lichman2013}. The goal of this study is to compare the models  DGD, HMGD, HMD1 \cite{togban2018classification}, DME  \cite{Togban2017}, Multinomial logistic regression (MLR) as well as the mixture of Generalized Dirichlet (MGD). The model HMD1 is a two levels hierarchical mixture model where the experts are multinomial logistic regressions. In HMD1, the first level gating function is based on Gaussian distribution and the second level gating function is chosen as a Softmax function. For the MGD model, each class is described by a Generalized Dirichlet and the class membership is given by the \textit{a posteriori} probability. Estimate the parameters of MGD is equivalent to estimate the parameters of DGD by setting the variational parameters to zero. The comparison between MGD and DGD will help us to evaluate the relevance of the upper-bound described above. Let us recall that HMD1 and DME combine several logistic regressions through different type of gating functions. In order to obtain compositional data we:1) delete the binary features, 2) standardize the data (mean is equal to zero and standard equal to one), 3) rescale the data between zero and one and  normalize them to get data lying in the simplex.  Table  (\ref{tab:Dataset-used-for_all_chap3}) present the main characteristics of the datasets used.

\captionsetup{width=1\columnwidth}
\begin{table}[!htb]
	\protect\caption{UCI Datasets used to evaluate the models \textit{DGD} and \textit{HMGD}. \label{tab:Dataset-used-for_all_chap3}}
	
	\begin{centering}
		\setlength 
		\setlength{\tabcolsep}{2pt}
		\def\arraystretch{1} 
		\begin{tabular*}{0.9\columnwidth}{@{\extracolsep{\fill}}lrrr}
			\textbf{Datasets} & \#instances & dimension & \#classes\tabularnewline
			\hline 
			
			Appendicitis & 106 & 7 & 2\tabularnewline
			Vowel & 990 & 10 & 11\tabularnewline
			Cancer (Breast)	&277&	9&	2\tabularnewline
			
			Magic&	19,020 &	10&	2\tabularnewline
			
			Vehicle&	946&	18&	4\tabularnewline
			Satimage&	6 435&	4&	6\tabularnewline
			
			\hline 
		\end{tabular*}
		\par\end{centering}

\end{table} 

\begin{figure}[!h]
	
	\begin{centering}
		\hfill{} \includegraphics[width=.8\columnwidth]{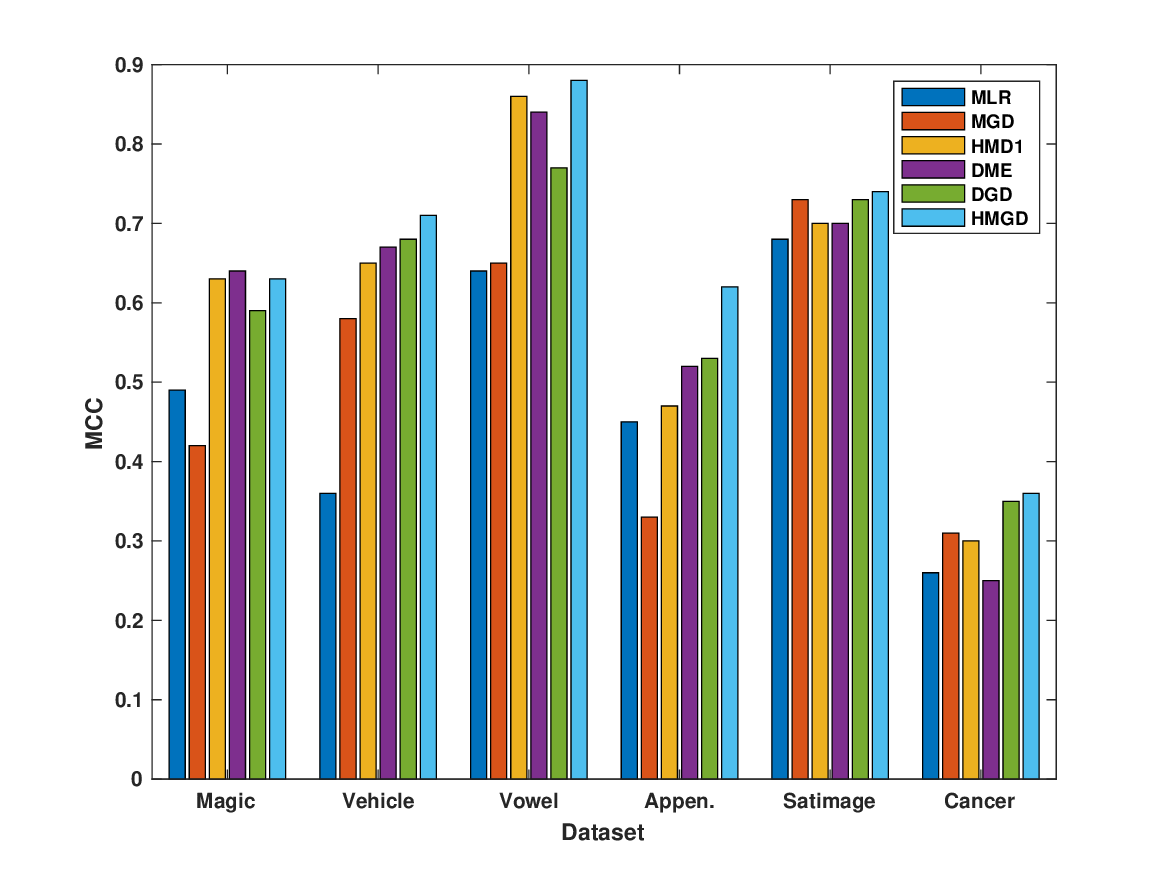}\hfill{}
		
		\protect\caption{Matthew Correlation Coefficient according to the results of table \ref{tab:Performance-vs-old}.\label{fig:mcc_exp1} }
	\end{centering}
\end{figure}

 The accuracies are reported in table   (\ref{tab:Performance-vs-old})  and the number in brackets refer to the number of experts used in  HMGD, HMD1, HMD2, DME. To choose the number of experts, we used the method described in \cite{togban2018classification}. We plot the MCC values in figure\ref{fig:mcc_exp1}.

\captionsetup{width=1\textwidth}
\begin{table*}[!htb]
		\protect\caption{ Comparison of DGD, MGD,  HMGD, DME, HMD1 and LR. \label{tab:Performance-vs-old}}
	\begin{centering}
		\setlength \extrarowheight{0pt} 
		\setlength{\tabcolsep}{2pt}
		\def\arraystretch{2}
		\resizebox{1\textwidth}{!}{%
			\begin{tabular}{|c|c|c|c|c|c|cIc|}
				\hline	
				\diagbox[width=9em]{\texttt{\textbf{Models}}}{\texttt{\textbf{Datasets}}} &  \makecell{\texttt{\textbf{Magic}} (5)}&
				\makecell{\texttt{\textbf{Veh.}}(4)}&
				\makecell{\texttt{\textbf{Vow.}} (4)}& \makecell{\texttt{\textbf{Appe.}} (2)}& \makecell{\texttt{\textbf{Sat.}} (4)}& \makecell{\texttt{\textbf{Cancer} (2)}}&
				\makecell{\texttt{\textbf{Mean}}}
				\tabularnewline
				\hline
				\makecell{MLR} & \makecell{78.5 \\ $\pm$ 0.39}&\makecell{40.8 \\ $\pm$  12.44  }&\makecell{64.04 \\ $\pm$  3.34 }&\makecell{84.89 \\ $\pm$  4.03 }& \makecell{74.33 \\ $\pm$  0.58  }&\makecell{72.94 \\ $\pm$  8.43 }& \makecell{69.25}
				\tabularnewline

				\hline
				\makecell{MGD} & \makecell{77.25 \\ $\pm$ 1.43}&\makecell{52.96 \\ $\pm$  5.6  }&\makecell{66.36 \\ $\pm$  4.47 }&\makecell{ 83.07\\ $\pm$ 5.11  }& \makecell{77.53 \\ $\pm$  0.69  }&\makecell{ 71.49\\ $\pm$ 4.6  }& \makecell{71.44}
				\tabularnewline

				\hline  
				\makecell{HMD1} &  \makecell{83.65 \\ $\pm$  0.74  }& \makecell{44.09 \\ $\pm$  6.56  }&\makecell{86.16 \\ $\pm$  4.53 }&\makecell{86.80 \\ $\pm$  2.08  }&\makecell{ 75.34\\ $\pm$ 0.62    }&\makecell{72.57\\ $\pm$ 6.93 }&\makecell{74.76}
				\tabularnewline
				\hline 
				
				\makecell{DME \cite{Togban2017}} &\makecell{\textbf{83.84} \\ $\pm$  0.53}&   \makecell{55.09 \\ $\pm$  6.5 }& \makecell{84.04 \\ $\pm$  3.81 }& \makecell{84.98 \\ $\pm$  8.26}& \makecell{75.31 \\ $\pm$  0.74 }&\makecell{70.4 \\ $\pm$  7.09 }& \makecell{75.61}
				\tabularnewline

				\hline 
				\hline
				\makecell{DGD} &  \makecell{82.23 \\ $\pm$ 0.8}&	\makecell{62.17 \\ $\pm$ 7.12}&	\makecell{79.49 \\ $\pm$ 3.08}&	\makecell{86.8 \\ $\pm$ 3.96}& \makecell{78.15 \\ $\pm$  0.61}& \makecell{72.22\\ $\pm$  4.6}&\makecell{76.52}
				
				\tabularnewline
				\hline
				\makecell{HMGD} & \makecell{83.22 \\ $\pm$ 0.83 }&\makecell{\textbf{68.91}\\ $\pm$ 8.37 }&\makecell{\textbf{88.79} \\ $\pm$ 0.9 }&\makecell{\textbf{87.75} \\ $\pm$ 2.5 }& \makecell{\textbf{78.91} \\ $\pm$  0.52 }& \makecell{\textbf{72.22} \\ $\pm$  3.09  }&\textbf{79.97}
				\tabularnewline
				\hline 
				
		\end{tabular}}
		
		\par\end{centering}

\end{table*}

\captionsetup[figure]{ width=0.9\columnwidth, justification=centering}

The results vary from one dataset to another. On Magic dataset,  DGD and HMGD  improve by at least  4\% the accuracy obtained by multinomial logistic regression and MGD. However, the standard deviations obtained by DGD and HMGD  are slightly higher than the one obtained by multinomial logistic regression. On this dataset,  HMD1 and DME slightly outperform HMGD (the variation of accruracy is below 1\%). With Vehicle dataset, DGD achieves an accuracy at least 7\% more than MGD, HMD1 and DME. However, the standard deviation of DGD is slighter higher than the ones obtained these four models (variation is below 1\%). The model DGD obtained a better accuracy compared to MGD on all the six datasets used in our experiment. In the case where the standard deviation obtained by MGD is lower than the one obtained by DGD (almost 2\% of variation), the accuracy of DGD is 9\% higher. The model DGD achieves a better accuracy compared to
HMD1 on the datasets   \textit{Vehicle} and \textit{Satellite}. With these two datasets, the standard deviation obtained by DGD  and HMD1 are similar.  The model HMGD achieves a better accuracy compared to HMD1 ( between 1\% and 6\%  higher) on datasets  Vowel, Appendicitis and Satimage. Moreover, on these three datasets, the standard deviation obtained by HMGD is lower than the one obtainded by HMD1. With the Cancer dataset, compared to HMD1, HMGD obtained a better standard deviation (3\% lower) while both achieve a similar accuracy. In general,  HMGD improve the accuracy and the standard deviation obtained by DGD. In the cases where the standard deviation obtained by HMGD is higher than the one obtained by DGD, the variation is small compared to the  gain in accuracy. For all datasets, the MCC values confirm with the scores obtained with the accuracy measure.

\subsection{Spam Detection}
Spams are irrelevant messages send to a set of recipients. These messages can contain malicious software or phishing programs.   According to  Kaspersky \cite{Kap2020}, 56.51 \% of the emails received in  2019 were spams. It is  important to detect those spams  in order to avoid boring the recipients, allowing them to focus on essential emails or protecting them from criminal activities.  For this purpose, several supervised methods have been used  \cite{jain2019optimizing,Ankam2018,dada2019machine}. We use our models DGD and HMGD to deal with spam detection problem. We compare our models against HMD1 \cite{togban2018classification},  MLR and  MGD. The goal here is to compare the strategy of applying the alpha-transformation (eq. \ref{eq:alpha-transf}) to the data with our models. Therefore, for the models HMD1 and MLR, we preprocessed the data with the $\alpha$-transformation (\ref{eq:alpha-transf}). In our experiments, we vary $\alpha$ from 0 to 1 with 0.2 as step. The accuracy is reported in table (\ref{tab:Performance-vs-spam}). The number in parentheses represents the number of experts used. The number following the symbol $\pm$ is the standard deviation.  We performed our experiments on two datasets previously used in the literature of spam detection \cite{kumar2012comparative,Ankam2018}.

i) \textbf{HP Spambase}  has been created by Hewlett-Packard Labs and downloaded from UCI machine learning repository.  This dataset has 4601 instances with 57  continuous attributes and the class label  (spams or not). Spams represent  39.4\% of the instances in this dataset. Since DGD and HMGD are designed for compositional data, we considered the 20 attributes representing the frequency of Words/Characters most used in this dataset. A spam detector can group the emails in three categories: legitimate, spams and needing more scrutiny. In order to create this third category, we randomly select 977 instances belonging to the legitimate category. Therefore, the new dataset has equal proportion of spams and legitimate messages (39.4\% each) and 21.2\% of messages needing more scrutiny. The accuracy  obtained by MLR and HMD1 are similar to the one obtained by DGD. However, HMGD improve the results of DGD and outperform HMD1. Let us recall that in this study, HMD1 combines two multinomial logistic regressions while HMGD combines two DGD.

ii) \textbf{Ling-spam} \footnote{\url{https://www.kaggle.com/mandygu/lingspam-dataset}} is a dataset containing  2893 emails with 481 (16.6\%) spams. We create a third category by selecting randomly 481 emails from the legitimate one. We obtain then a dataset composed of 1443 emails (481 emails per category). The features used are obtained by first, deleting stop words and by doing a lemmatization \cite{Ankam2018}. After that, a word dictionary is created with the 30 most common words. The final feature is a vector containing the words frequency. 

The models  DGD and HMGD outperformed  MLR, HMD1 and MGD in terms of accuracy (with lower standard deviation in the case of HMGD). We notice that HMGD improve both the accuracy  and standard deviation obtained by DGD. The models  DGD and HMGD achieve an accuracy of  2\% to 3\%  higher than HMD1 and MLR  using $\alpha$-transformation. As noted in the previous experiments, the variational approximation apply to the model DGD allow us to achieve better results compared to MGD. The MCC values obtained by DGD and HMGD suggest that the two models achieve a slightly better classification than MLR, MGD and HMGD.
\captionsetup{width=1\columnwidth}
\begin{table*}[!tbh]
		\protect\caption{Comparison of DGD, HMGD, HMD1, MLR  and MGD on the datasets HP Spambase and Ling-spam. \label{tab:Performance-vs-spam}}
	\centering
	\addtolength\tabcolsep{2pt}
	\def\arraystretch{2}
	\begin{tabular}{|c|c|c|c|c|c|}
		
		\cline{2-6}
		\multicolumn{1}{c|}{}&	\makecell{MLR}& MGD& \makecell{HMD1  \\ (2)}&DGD&\makecell{HMGD \\ (2)}
		\tabularnewline
		
		\hline
		
		HP Spambase&\makecell{73.51  \\$\pm$ 0.9 \\($\mathbf{\alpha}^{*}=0$)}&\makecell{71.30  \\ $\pm$  0.83}&\makecell{73.49 \\ $\pm$ 0.66\\($\mathbf{\alpha}^{*}=0$)}&\makecell{73.60  \\ $\pm$ 0.85}&\makecell{\textbf{74.47}\\$\pm$ 0.72}
		\tabularnewline
		\hline
		\hline
		
		Ling-spam&\makecell{62.81  \\$\pm$ 2.69 \\($\mathbf{\alpha}^{*}=0.6$)}  &\makecell{63.93 \\ $\pm$  2.92}&\makecell{62.67 \\ $\pm$ 2.56\\($\mathbf{\alpha}^{*}=0.6$)}&\makecell{64.81 \\ $\pm$ 3.01}&\makecell{\textbf{66.07} \\ $\pm$ 2.10}
		\tabularnewline
		\hline

	\end{tabular}

\end{table*}  

\captionsetup[figure]{ width=0.8\columnwidth, justification=centering}
\begin{figure}[!h]
	
	\begin{centering}
		\hfill{} \includegraphics[width=1\columnwidth]{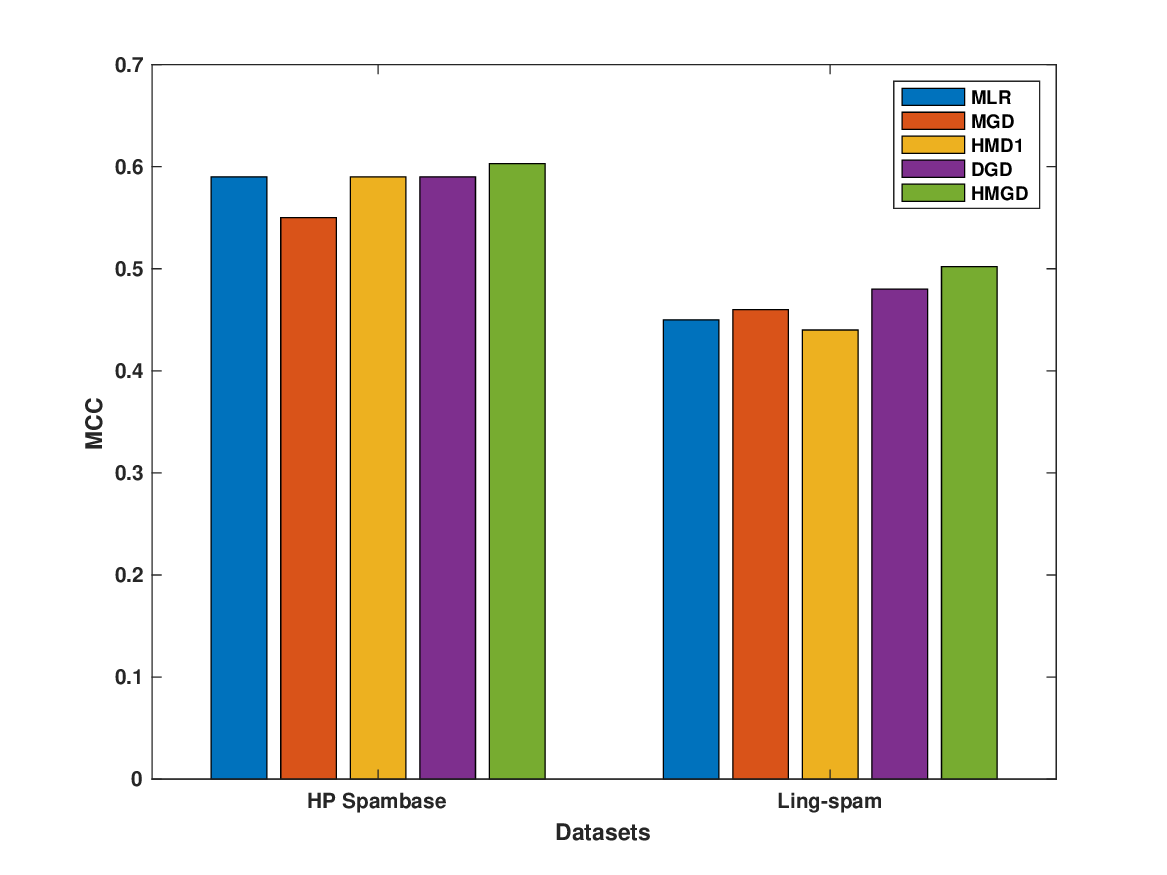}\hfill{}
		
		\protect\caption{Matthew Correlation Coefficient according to the results of table \ref{tab:Performance-vs-spam}.\label{fig:mcc_exp2} }
	\end{centering}
\end{figure}


\subsection{color space identification}
Color space is a color representation system based on a set of triplets.  One of the most used color space for  visualization is RGB, which is composed of the channels Red, Green and Blue. In order to faithfully represent color in different human activities (photography, displaying, industry), several variants of RGB have emerged such as sRGB, Adobe RGB, Apple RGB and ProPhoto RGB. It is important to correctly choose the color space of an image if we want to faithfully reproduce it. Some applications choose the color space by default for the purpose of displaying. For example, the color space use by default on the web is sRGB while image-editing software Adobe Lightroom use  Adobe RGB.  After editing an image, it is possible to embed in EXIF data, the information relative to the color space. Once this information is available, one can faithfully represent the image with any displaying devices. But for different reasons, the Exif data can be missing or damaged which will lead to a misrepresentation of the image.

The goal of our experiment is to discriminate between the color space in the RGB family. For this experiment, we have considered the color spaces sRGB, Apple RGB, ColorMatch RGB, Adobe RGB  and ProPhoto RGB. The problem of color space identification was first addressed by  Vezina et al. \cite{martin2015}. In their work, they used gamut estimation to discriminate between sRGB, HSV, HLS and Lab color spaces. Unfortunately, in the case of the RGB family, we need a more accurate estimation of the gamut. However, it is possible to exploit the correlation between the pixels  for color space identification. Let $I$ be a color image in a RGB color space. We assume that if there is a correlation between the pixels, their value can be expressed as  follows:

\begin{equation}
	\begin{array}{r}
		\thinmuskip=2mu\medmuskip=2mu\thickmuskip=2mu
		I(m,n,k)=\underset{ \makecell {(i,j) \; \in  V(m, n)} }{\sum}\gamma_{i,j,k}I(m+i,n+j,k)+w(m,n)

	\end{array}
	\label{eq: demosai_equa}
\end{equation}
where $\gamma_{i,j,k}$  are a set of coefficients with $\gamma_{0,0,k}=0$,  $k \in \{R,G,B\}$ refer to the channels of RGB, $I(m,n,k)$ is the value of the pixel located at $(m,n)$ for the channel $k$ and $V(m, n)$ its  neighborhood. The variable $w(m,n)$ is a Gaussian white noise with standard deviation $\sigma$.  Basically, we assume that if the pixels are correlated, each pixel value  can be expressed as a linear function of its adjacent pixels. The parameters of the model described in equation \ref{eq: demosai_equa} are given by the vector  $\gamma_{k}$ with $\{\gamma_{i,j,k}\mid (i,j) \in V(m, n)\}$ as elements and $\sigma$. In this paper, we define the neighborhood as $V(m,n)= \{(i,j)\mid -J\leq i,j \leq J\}$ where $J$ denotes the number of pixels located on either side of the pixel at the location $(m,n)$. In a recent work \cite{color_2023}, we show that the correlation coefficients $\gamma_{i,j,k}$ can be used as features to identify RGB color space. The features dimension is  $3(2J+1)^{2}-3$.

\captionsetup{width=1\columnwidth}
\begin{table}[!tbh]
		\protect\caption{Accuracy obtained by  DGD, HMGD, HMD1, MLR and MGD  for the color space identification problem. \label{tab:Performance-vs-color}}
	\centering
	\setlength \extrarowheight{0pt} 
	\setlength{\tabcolsep}{2pt}
	\def\arraystretch{2}
	\begin{tabular}{ccc|c|c|}
		\tabularnewline
		\cline{2-5}
		\multicolumn{1}{ c|  }{}&\multicolumn{1}{ c|  }{$\mathbf{\alpha}^{*}$ }&\makecell{\textbf{Color}\\\textbf{Space}}&\makecell{\textbf{Accuracy by}\\\textbf{ space (\%)}} &\makecell{\textbf{Accuracy}}
		\tabularnewline
		\hline  
		\multicolumn{1}{ |c|  }{MLR}&\multicolumn{1}{ c|  }{0}&\makecell{\small{Adobe}  \\ Apple \\ColorMatch \\ ProPhoto\\ sRVB} &\makecell{ 63.25 \\ 63.85 \\ 78.5 \\ 67.2  \\ 61.6 }& 66.88 $\pm$ 1.17
		\tabularnewline
		\hline
		\multicolumn{1}{ |c|  }{HMD1 }&\multicolumn{1}{ c|  }{1}&\makecell{Adobe  \\ Apple \\ColorMatch \\ ProPhoto\\ sRVB} &\makecell{ 62.25 \\62.59 \\\textbf{82.4} \\\textbf{70.3} \\ 61}& 67.71 $\pm$ 0.91
		\tabularnewline
		\hline
		\multicolumn{2}{ |c|  }{MGD}&\makecell{Adobe  \\ Apple \\ColorMatch \\ ProPhoto\\ sRVB} &\makecell{ \textbf{82.9} \\ 61.54  \\ 63.65\\62.7  \\60.4}& 66.24 $\pm$ 1.40
		\tabularnewline
		\hline
		\multicolumn{2}{ |c|  }{DGD}&\makecell{Adobe  \\ Apple \\ColorMatch \\ ProPhoto\\ sRVB} &\makecell{ 72 \\ 60.14 \\ 78.05\\ 66.55 \\ 63.5}& 68.05 $\pm$ 1.19
		\tabularnewline
		\hline
		\multicolumn{2}{ |c|  }{HMGD}&\makecell{Adobe  \\ Apple \\ColorMatch \\ ProPhoto\\ sRVB} &\makecell{ 70.55 \\ \textbf{67.75} \\ 78.05 \\ 68 \\ 63}& \textbf{69.58} $\pm$ 0.75
		\tabularnewline
		\hline

	\end{tabular}

\end{table}

We collect  1000  images  in RAW format from each  DRESDEN \cite{gloe2010dresden} and RAISE \cite{dang2015raise} collections. Then, we convert each  image in each of the five color spaces (10K images in all). In order to reduce the computational cost, the neighborhood is chosen by setting $J=2$. Once the  coefficients $\gamma_{i,j,k}$ have been extracted, we use the Generalized discriminant analysis to reduce the dimensions of the features \cite{baudat2000generalized}. Our models (DGD and HMGD) are compared with HMD1, MLR and MGD. We use two experts in HMD1 and HMGD. We apply the $\alpha$-transformation to the data before classifying them with MLR and HMD1. The accuracy (in percentage) is reported in table \ref{tab:Performance-vs-color}. The accuracy score here is adequate since the classes are well balanced.

HMGD and DGD outperform all others methods in terms of accuracy. However the accuracy obtained by DGD is sligthly better than the one obtained by HMD1 (less than 1\%). HMGD improve the accuracy obtained by DGD and reduce the standard deviation. We notice that HMD1 and HMGD have a  standard deviation lower than the one obtained by other methods. In general, HMD1 is better for the identification of ColorMatch RGB and Prophoto RGB. The model MGD  and HMGD are respectively better for the identification of Adobe RGB and Apple RGB.

\subsection{Limitations}
  
 Even though our models can improve the classification of compositional data and we don't need lot of data for training, they have some limitations. First,
 the computational complexity of DGD depends on the number of EM iterations and the computation of the parameters. The computational complexity of each iteration is around $\mathcal{O}(D^2 +(C-1)^2 + N)$ where $D$ is the dimension, $C$ the number of classes and $N$ the number of training instance. The calculation cost increases linearly with the number of data and quadratically with the dimension and number of classes. Fortunately, the first attribute is the one that can be quite large compared to the other two. As the number of training instances increases, the learning process could be accelerated by parallelizing the calculation of variational parameters. In the case of HMGD, in addition to $N, C, D$, the computational complexity depends on the number of experts $N_e$, the top as well as inner branching factor ($K$ and $M$), $\mathcal{O}(D^2 +(C-1)^2 + N)N_eKM$. Here, we supposed that $M_i \equiv M$.  Second, in some case, the HMGD does not improve the accuracy of DGD (e.g. for Cancer dataset). This can happen when we are in presence of small dataset. This limitation could be overcome by using  prior distribution over parameters. This type of Bayesian approach has shown improvement in the inference of compositional data \cite{fan2017proportional}. Third, We also suspect that  it can be difficult to find suitable regions  when the  classes are highly imbalanced, in such case the experts cannot correctly discriminate these classes. This limitation could be overcome by adding weights in the model. In order to reduce the overfitting, one can use a prior distribution for the parameters of HMGD.

\section{Conclusion\label{Conclusion}}
In this paper, we deal with the supervised classification of compositional data. For this purpose, we used a posterior probability based on Generalized Dirichlet distribution as model (DGD). Moreover,  we proposed a hierarchical mixture of DGD experts (HMGD). Through a variational approximation, we estimated the parameters of DGD and HMGD. This approximation is possible thanks to the derivation of an upper-bound for the Generalized Dirichlet mixture. At the best of our knowledge, it is the first time this bound is proposed in the literature. Experiments with spam detection and color space identification suggest that DGD and HMGD can be alternatives to the use of $\alpha$-transformation. Our findings suggest that in discriminative learning, the compositional data can be modeled in a simplex.   Finally, we noted that in certain cases, HMGD do not improve the predictions of DGD. This is be due to overfitting and could be overcome by adding some constraints on HMGD. Moreover, with some datasets, it is hard to find several regions in which we have  all the classes. In this case, some experts perform poorly which leads to an increase of the classification errors.

\appendix 
\section{Upper-bound for the GD mixture model \label{appendix_upper}}
For more insights on the derivation of this upper-bound, readers can refer to the works of Jebara et al. \cite{jebara2001discriminative}. In the following lines, we applied the results obtained in \cite{jebara2001discriminative} for the specific case of the GD mixture model.

\section*{Re-parametrization of GD mixture model}
The Gd mixture can be re-parametrized as a sum of exponential distribution and the formulation is as follows:
\begin{equation}
	\overset{M_i}{\underset{k=1}{\sum}}\alpha_{ik}^{(l)}GD(x_{n}|\mu_{ik}^{(l)})= \overset{M_i}{\underset{k=1}{\sum}}\exp\left[ \bar{\mathbf{x}}_{n,ik}^{(l)T}\bar{\Omega}_{ik}^{(l)}-\mathcal{K}(\bar{\Omega}_{ik}^{(l)}) \right] \label{re-expo}
\end{equation}
where
\begin{equation*}
	\begin{array}{lll}
		\bar{\mathbf{x}}_{n,ik}^{(l)}=
		\begin{pmatrix}
			\log(\mathtt{v}_{n,1})\\
			\log(A-\mathtt{v}_{n,1})\\
			\vdots\\
			\log(\mathtt{v}_{n,D})\\
			\log(A-\mathtt{v}_{n,D})\\
			\dot{\mathbf{x}}_{k}
		\end{pmatrix}; &\bar{\Omega}_{ik}^{(l)}=
		\begin{pmatrix}
			a_{ik,1}^{(l)}\\
			b_{ik,1}^{(l)}\\
			\vdots\\
			a_{ik,D}^{(l)}\\
			b_{ik,D}^{(l)}\\
			\eta_{i}^{(l)}
		\end{pmatrix};
		&
		\eta_{i}^{(l)}=
		\begin{pmatrix}
			\eta_{i,1}^{(l)}\\
			\vdots\\
			\eta_{i,M_i-1}^{(l)}
		\end{pmatrix}
	\end{array}
\end{equation*}

\begin{equation*}
	\begin{array}{c}
		\mathcal{K}(\bar{\Omega}_{ik}^{(l)})=\ln\left(1+\overset{M_i-1}{\underset{k=1}{\sum}}\exp\left(\eta_{i,k}^{(l)} \right)  \right) + \overset{D}{\underset{d=1}{\sum}}\left[\ln \Gamma\left( a_{ik,d}^{(l)}\right) +    \right. \\ \left. \left( a_{ik,d}^{(l)}+b_{ik,d}^{(l)}-1\right)\ln(A)+  \ln\Gamma\left( b_{ik,d}^{(l)}\right)+ \ln \Gamma\left( a_{ik,d}^{(l)}+b_{ik,d}^{(l)}\right)\right] \\
	 \eta_{i,k}^{(l)}= \mbox{\Large\( \ln\left(\sfrac{\alpha_{i,k}^{(l)}}{\alpha_{i,M_i}^{(l)}} \right) \)}
	\end{array}
\end{equation*}
with $\Gamma(.)$ is the Gamma function and $\dot{\mathbf{x}}_{k}$ is a vector of dimension $(M_i-1)$ where the $k^{th}$ element is equal to $1$ and the remaining  are $0$. Given the expression in equation (\ref{re-expo}), we can derive the upper bound of the GD mixture (Eq. \ref{eq: upper-bound}) as shown in \cite{jebara2001discriminative} for the exponential family in general. The first derivative of the cumulant function is given by the following  vector of dimension  $(2D+M_i)$:
\vspace{-.5cm}
\begin{equation}
	\mathcal{K}^{\prime}(\bar{\Omega}_{ik}^{(l)})=
	\begin{pmatrix}
		\mathsf{\Psi} (\mu_{ik,1}^{(l)})-\mathsf{\Psi} (|\mu_{ik,1}^{(l)}|)+\ln(A)\\
		\vdots \\
		\mathsf{\Psi} (\mu_{ik,D}^{(l)})-\mathsf{\Psi} (|\mu_{ik,D}^{(l)}|)+\ln(A)\\
		
		\sfrac{ \left( \exp \left(\eta_{i}^{(l)} \right) \right)}{ \overset{M_i}{\underset{j=1}{\sum}}\exp\left(\eta_{i,j}^{(l)} \right)}
	\end{pmatrix}
\end{equation}
The Hessian of the cumulant function is a diagonal matrix  per bloc:
\vspace{-.5cm}
\begin{equation}
	\mathcal{K}^{\prime \prime}(\bar{\Omega}_{ik}^{(l)})=
	\begin{pmatrix}
		\mathcal{K}_{1}^{\prime \prime}(\bar{\Omega}_{ik}^{(l)})& \mathbf{0} \\
		\mathbf{0}&\mathcal{K}_{2}^{\prime \prime}(\bar{\Omega}_{ik}^{(l)})
	\end{pmatrix}
\end{equation}
where $\mathcal{K}_{1}^{\prime \prime}(\bar{\Omega}_{ik}^{(l)})$ is a  diagonal matrix per bloc composed by $D$ blocs

\vspace{-1cm}

\begin{equation*}
	\renewcommand{\arraystretch}{2}
	\begin{array}{l}
		\mathcal{K}_{1,d}^{\prime \prime}(\bar{\Omega}_{ik}^{(l)})= 
		\begin{pmatrix}
			\mathsf{\Psi}' (a_{ik,d}^{(l)})-\mathsf{\Psi}' (|\mu_{ik,d}^{(l)}|)&-\mathsf{\Psi}' (|\mu_{ik,d}^{(l)}|)\\
			-\mathsf{\Psi}' (|\mu_{ik,d}^{(l)}|)& \mathsf{\Psi}'(b_{ik,d}^{(l)})-\mathsf{\Psi}' (|\mu_{ik,d}^{(l)}|)
		\end{pmatrix}
	\end{array}
\end{equation*}
The matrix $\mathcal{K}_{2}^{\prime \prime}(\bar{\Omega}_{ik}^{(l)})$ elements  are given for $d_1$ and  $d_2 \in \left\lbrace 1,\dots,M_i-1\right\rbrace$ by:
\begin{equation}
	\begin{array}{l}
		\mathcal{K}_{2,d_1d_2}^{\prime \prime}(\bar{\Omega}_{ik}^{(l)})= 
		\begin{cases}
			\mathcal{K}_{(2D+d_1)}^{\prime}(\bar{\Omega}_{ik}^{(l)}) \left[1-\mathcal{K}_{(2D+d_1)}^{\prime}(\bar{\Omega}_{ik}^{(l)})\right], & \text{if } d_1=d_2 \\ 
			-\mathcal{K}_{(2D+d_1)}^{\prime}(\bar{\Omega}_{ik}^{(l)})\mathcal{K}_{(2D+d_2)}^{\prime}(\bar{\Omega}_{ik}^{(l)}),& \text{if } d_1 \neq d_2
		\end{cases}
	\end{array}
	\label{eq:k2}
\end{equation}

\section*{Constraints on $\ddot{\mathbf{x}}_{n,ik}^{(l)}$ }

The variational point $\ddot{\mathbf{x}}_{n,ik}^{(l)}$ has to belong to the tangent space generated by $\mathcal{K}(\bar{\Omega}_{ik}^{(l)})$. For that, we have to compute $w_{n,ik}^{(l)}$ as the minimum value of $W_{n,ik}^{(l)}$ that guarantee this constraint on $\ddot{\mathbf{x}}_{n,ij}^{(l)}$. Let us recall that $\bar{\mathbf{x}}_{n,ik}^{(l)}$ belong to the gradient space of $\mathcal{K}(\bar{\Omega}_{ik}^{(l)})$. All we need is to apply the constraints of $\bar{\mathbf{x}}_{n,ik}^{(l)}$ to $\ddot{\mathbf{x}}_{n,ik}^{(l)}$. The following conditions must be satisfied:
\begin{itemize}
	\item[\textbullet] the first $2D$ elements of  $\ddot{\mathbf{x}}_{n,ik}^{(l)}$ must be smaller than $\ln (A)$
	\item[\textbullet] the remaining elements must be in the interval $]0,1[$ and their sum must be equal to one.	
\end{itemize}
Without loss of generality, $w_{n,ik}^{(l)}$ has to satisfy the following constraint:

\begin{equation}
	w_{n,ik}^{(l)}>\hat{B}_{n,ik}^{(l)}=\max
	\begin{Bmatrix}
		B_{n,ik,1}^{(l)};\cdots;B_{n,ik,(2D+2M_i)}^{(l)} 
	\end{Bmatrix}
\end{equation}

\begin{equation*}
	B_{n,ik,d}^{(l)}=
	\begin{cases}
		\text{for } d=1 \cdots 2D \\
		
		\tilde{\pi}_{k|i}^{(l)}\left(x_{n}\right) \dfrac{\left[ \bar{\mathbf{x}}_{n,ik,d}^{(l)}- \mathcal{K}_{d}^{\prime}(\breve{\Omega}_{ik}^{(l)}) \right]}{ \mathcal{K}_{d}^{\prime}(\breve{\Omega}_{ik}^{(l)})-\ln(A)}; \\ 
		\text{for } d=2D+1 \cdots 2D+M_i-1
		\\
		\tilde{\pi}_{k|i}^{(l)}\left(x_{n}\right) \dfrac{\left[ \bar{\mathbf{x}}_{n,ik,d}^{(l)}- \mathcal{K}_{d}^{\prime}(\breve{\Omega}_{ik}^{(l)}) \right]}{ \mathcal{K}_{d}^{\prime}(\breve{\Omega}_{ik}^{(l)})};
		
		\\  
		\text{for } d=2D+M_i \cdots 2D+2M_i-1
		\\
		\tilde{\pi}_{k|i}^{(l)}\left(x_{n}\right) \dfrac{\left[ \bar{\mathbf{x}}_{n,ik,(d-M_i+1)}^{(l)}- \mathcal{K}_{(d-M_i+1)}^{\prime}(\breve{\Omega}_{ik}^{(l)}) \right]}{ \mathcal{K}_{(d-M_i+1)}^{\prime}(\breve{\Omega}_{ik}^{(l)})-1} \\ 
		\text{for } d=2D+2M_i \\
		
		\tilde{\pi}_{k|i}^{(l)}\left(x_{n}\right) \dfrac{ \overset{2D+M_i-1}{\underset{j=2D+1}{\sum}} \left[ \bar{\mathbf{x}}_{n,ik,j}^{(l)}- \mathcal{K}_{j}^{\prime}(\breve{\Omega}_{ik}^{(l)}) \right]}{ \overset{2D+M_i-1}{\underset{j=2D+1}{\sum}} \mathcal{K}_{j}^{\prime}(\breve{\Omega}_{ik}^{(l)})-1}
	\end{cases}
\end{equation*}

\section*{Inversion of $\mathcal{K}^{\prime \prime}(\breve{\Omega}_{ik}^{(l)})$ and $\mathcal{H}_{ik,d}^{(l)}$ \label{appendix_inv}}
We need to invert $\mathcal{K}^{\prime \prime}(\breve{\Omega}_{ik}^{(l)})$ and $\mathcal{H}_{ik,d}^{(l)}$ several times in our model. This can lead to a heavy computational and storage burden and avoid this will be a benefit. To invert $\mathcal{K}^{\prime \prime}(\breve{\Omega}_{ik}^{(l)})$ is equivalent to the inversion of each $\mathcal{K}_{1,d}^{\prime \prime}(\breve{\Omega}_{ik}^{(l)})$ and $\mathcal{K}_{2}^{\prime \prime}(\breve{\Omega}_{ik}^{(l)})$. Let us rewrite each one of these matrices as follows:

\begin{equation*}
	\begin{array}{l}
		
		\mathcal{K}_{1,d}^{\prime \prime}(\breve{\Omega}_{ik}^{(l)})=Q_{1}+e_{1} \mathbf{1} \mathbf{1}^{T} \; \mid e_1=-\mathsf{\Psi}' (|\mu_{ij,d}^{(l)}|) \\
		
		\mathcal{K}_{2}^{\prime \prime}(\breve{\Omega}_{ik}^{(l)})=Q_{2}+e_{2}\mathtt{\mathbf{f}}  \mathtt{\mathbf{f}}^{T} \; \mid e_2=-1\\
		
		\mathcal{H}_{ik,d}^{(l)}=R+e_{3} \mathtt{g} \mathtt{g}^{T} \; \mid \mathtt{g}^{T}=(a_{ik,d}^{(l)},b_{ik,d}^{(l)})\\
		 e_3=\mathsf{\Psi}' (|\mu_{ij,d}^{(l)}|)\underset{n}{\sum}H_{n,i}^{(l-1)}
		\left[h_{n,k|i}^{(l)}+W_{n,ik}^{(l)}\right]
	\end{array}
\end{equation*}
with $Q_{1}$ is a $2\times2$ diagonal matrix with entries $\mathsf{\Psi}' (a_{ik,d}^{(l)})$ and  $\mathsf{\Psi}' (b_{ik,d}^{(l)})$.

$\mathbf{1}$ is a column vector filled with 1, $Q_{2}$ and $\mathtt{\mathbf{f}}$ have the same elements  $\left\{\mathcal{K}_{d}^{\prime}(\bar{\Omega}_{ik}^{(l)})\right\}_{d=2D+1}^{2D+M_i-1} $ but the former is a diagonal matrix and the latter is a column vector. The matrix $R$ is a $2\times2$ diagonal matrix with entries: 
\vspace{-.5cm}
\begin{equation*}
	\begin{array}{rl}
		\begin{pmatrix}
		 
			R_1
			\\
			
			\\
			R_2
		\end{pmatrix}=\begin{pmatrix}
			-[a_{ij,d}^{(l)}]^{2}\mathsf{\Psi}' (a_{ij,d}^{(l)}) \underset{n}{\sum}H_{n,i}^{(l-1)} \left(h_{n,j|i}^{(l)}+W_{n,ij}^{(l)}\right) 
			+\left[ \dfrac{\partial \; \Phi_{1}^{(l)}}{\partial\xi_{ij,d}^{(l)}}\right]_{1}
			\\

			\\
			-[b_{ij,d}^{(l)}]^{2}\mathsf{\Psi}' (b_{ij,d}^{(l)}) \underset{n}{\sum}H_{n,i}^{(l-1)} \left(h_{n,j|i}^{(l)} +W_{n,ij}^{(l)}\right) + \left[ \dfrac{\partial \; \Phi_{1}^{(l)}}{\partial\xi_{ij,d}^{(l)}}\right]_{2}
		\end{pmatrix}
	\end{array}
\end{equation*}
where  $\left[ \dfrac{\partial \; \Phi_{1}^{(l)}}{\partial\xi_{ij,d}^{(l)}}\right]_{1}$ and $\left[ \dfrac{\partial \; \Phi_{1}^{(l)}}{\partial\xi_{ij,d}^{(l)}}\right]_{2}$ are respectively the  first and  second components of $ \dfrac{\partial \; \Phi_{1}^{(l)}}{\partial\xi_{ij,d}^{(l)}}$. Using the Sherman-Morrison formula, the inverses can be written as follows:
\begin{equation*}
	\begin{array}{rl}
		\mathcal{K}_{1,d}^{\prime \prime}(\breve{\Omega}_{ik}^{(l)})^{-1}=Q_{1}^{-1}-e_{1}^{*} u u^{T} \; \mid u^{-1}= \begin{pmatrix}
			\mathsf{\Psi}' (a_{ik,d}^{(l)})\\
			
			\mathsf{\Psi}' (b_{ik,d}^{(l)})
		\end{pmatrix}
	\end{array}
\end{equation*}
\vspace{-.5cm}
\begin{equation*}
	\begin{array}{rl}
		\mathcal{K}_{2}^{\prime \prime}(\breve{\Omega}_{ik}^{(l)})^{-1}=Q_{2}^{-1}+e_{2}^{*}\mathbf{1} \mathbf{1}^{T} \; \mid \mathlarger{\frac{1}{e_{2}^{*}}}=1-\overset{2D+M_i-1}{\underset{d=2D+1}{\sum}} \mathcal{K}_{d}^{\prime}(\bar{\Omega}_{ik}^{(l)})
	\end{array}
\end{equation*}
\vspace{-.5cm}
\begin{equation*}
	\begin{array}{rl}
		
		[\mathcal{H}_{ik,d}^{(l)}]^{-1}=R^{-1}-e_{3}^{*} \mathtt{g}^{*} \mathtt{g}^{*T} \; \mid \mathtt{g}^{*T}=\begin{pmatrix}
			\mathlarger{\frac{a_{ik,d}^{(l)}}{R_1}}\; \; ,			\mathlarger{\frac{b_{ik,d}^{(l)}}{R_2}}
		\end{pmatrix} 
	\end{array}
\end{equation*}
\vspace{-.5cm}
\begin{equation*}
	\begin{array}{rl}
		
		e_3= e_3^{*}\big[1+e_{3}\mbox{\Large\(\big[\frac{[a_{ij,d}^{(l)}]^{2}}{R_1}+\frac{[b_{ij,d}^{(l)}]^{2}}{R_2}\big] \)}\big]  
		
	\end{array}
\end{equation*}
\vspace{-.5cm}
\begin{equation*}
	\begin{array}{rl}
		
		e_{1}=  e_{1}^{*}\big[1+e_{1} \mbox{\Large\([\frac{1}{\mathsf{\Psi}' (a_{ik,d}^{(l)})}+\frac{1}{\mathsf{\Psi}' (b_{ik,d}^{(l)})}]\)} \big].
		
	\end{array}
\end{equation*}.

\vspace{-.5cm}

\bibliographystyle{elsarticle-num}
\bibliography{HMD_dir_jrnl_elsevier}

\vspace{1.5cm}
\noindent{ \bf Elvis Togban} received the B.E. degree in Telecommunication from {\'E}cole Sup{\'e}rieure des communications de Tunis (Sup'Com). He received a Ph.D. degree at the Universit{\'e} de Sherbrooke, Canada in  2021.  His current research interests include machine
	learning, data mining, and pattern recognition. 

\vspace{1.5cm}

\noindent{ \bf Djemel Ziou} received the B.E. degree in computer science from the University of Annaba, Annaba,
	Algeria, in 1984, and the Ph.D. degree in computer science from the Institut National Polytechnique de
	Lorraine, Nancy, France, in 1991.   He served as a Lecturer in several universities in France from 1987 to 1993. He is currently a Full Professor with the Department of Computer Science, Universit{\'e} de Sherbrooke, Sherbrooke, QC, Canada. He holds the Natural Sciences and Engineering Research Council/Bell Canada Research Chair in personal imaging. His current research interests include image processing, information retrieval, computer vision, and pattern recognition. Dr. Ziou has served on numerous conference committees as member or
	chair. He heads the Laboratory MOIVRE and the consortium CoRIMedia, which he founded.

\end{document}